\documentclass[letterpaper, 10 pt, conference]{ieeeconf}
\IEEEoverridecommandlockouts 

\usepackage{cite}
\usepackage{amssymb}
\usepackage{amsmath}
\usepackage{amsfonts}
\DeclareMathOperator*{\argmin}{arg\,min}
\usepackage{algorithmic}
\usepackage[linesnumbered,ruled,vlined]{algorithm2e}
\usepackage{algorithm2e}
\usepackage{graphicx}
\usepackage{hyperref}
\usepackage[nameinlink]{cleveref}
\usepackage[font={small}]{caption}
\usepackage{textcomp}
\usepackage{subfigure}
\usepackage{xcolor}
\usepackage{multirow}

\def\BibTeX{{\rm B\kern-.05em{\sc i\kern-.025em b}\kern-.08em
    T\kern-.1667em\lower.7ex\hbox{E}\kern-.125emX}}
\renewcommand{\footnoterule}{\hrule width \columnwidth height 0.4pt \kern 5pt}

\newtheorem{remark}{Remark}
\usepackage{xcolor}

\begin{document}

\title{ Geometric Projectors: Geometric Constraints based Optimization for Robot Behaviors

\thanks{
}}

\author{
Xuemin Chi$^{1,2}$, Tobias Löw$^{2,3}$, Yiming Li$^{2,3}$, Zhitao Liu$^{1,*}$, Sylvain Calinon$^{2,3,*}$
\thanks{$^1$ Zhejiang University, Hangzhou, CN.}
\thanks{\texttt{\{chixuemin,ztliu\}@zju.edu.cn}}
\thanks{$^2$ Idiap Research Institute, Martigny, CH.}
\thanks{\texttt{name.surname@idiap.ch}}
\thanks{$^3$ Ecole Polytechnique Fédérale de Lausanne (EPFL), CH.}
\thanks{$^*$ Corresponding author.}
}
\newcommand{\trsp}{{\scriptscriptstyle\top}}
\newcommand{\psin}{{\dagger}}
\newcommand{\tp}[1]{\text{\tiny#1}}
\newcommand{\ty}[1]{{\scriptscriptstyle{#1}}}
\newcommand{\diag}{\mathrm{diag}}
\newcommand{\tr}{\mathrm{tr}}
\newcommand{\cov}{\mathrm{cov}}

\maketitle

\begin{abstract}
Generating motion for robots that interact with objects of various shapes is a complex challenge, further complicated when the robot’s own geometry and multiple desired behaviors are considered. 
To address this issue, we introduce a new framework based on Geometric Projectors (GeoPro) for constrained optimization.
This novel framework allows for the generation of task-agnostic behaviors that are compliant with geometric constraints. 
GeoPro streamlines the design of behaviors in both task and configuration spaces, offering diverse functionalities such as collision avoidance and goal-reaching, while maintaining high computational efficiency. 
We validate the efficacy of our work through simulations and Franka Emika robotic experiments, comparing its performance against state-of-the-art methodologies. 
This comprehensive evaluation highlights GeoPro's versatility in accommodating robots with varying dynamics and precise geometric shapes.
For additional materials, please visit: \href{https://www.xueminchi.com/publications/geopro}{https://www.xueminchi.com/publications/geopro}

\end{abstract}
\section{introduction}
In this paper, we focus on the generation of composable, non-conservative, and smooth behaviors for robots, particularly addressing three categories of behaviors: collision avoidance, goal-reaching, and self-limitation. Collision avoidance is a ubiquitous requirement in robotic applications and demands precise geometric modeling of both the robots and their surrounding environments for effective, non-conservative solutions.
The use of Signed Distance Functions (SDFs) is prevalent in this context but introduces challenges in the problem formulation. Traditional approaches like TrajOpt~\cite{schulman2014motion} have ignored non-differentiable points in configuration space, justifying this omission empirically. However, recent advancements in dual approaches~\cite{zhang2020optimization} have successfully transformed non-differentiable SDFs into twice-differentiable constraints. While this offers a theoretical solution, the computational complexity of this method limits its applicability to high-dimensional systems or intricate environmental scenarios.
\begin{figure}[t!]
    \centering
    \includegraphics[width=0.4\textwidth]{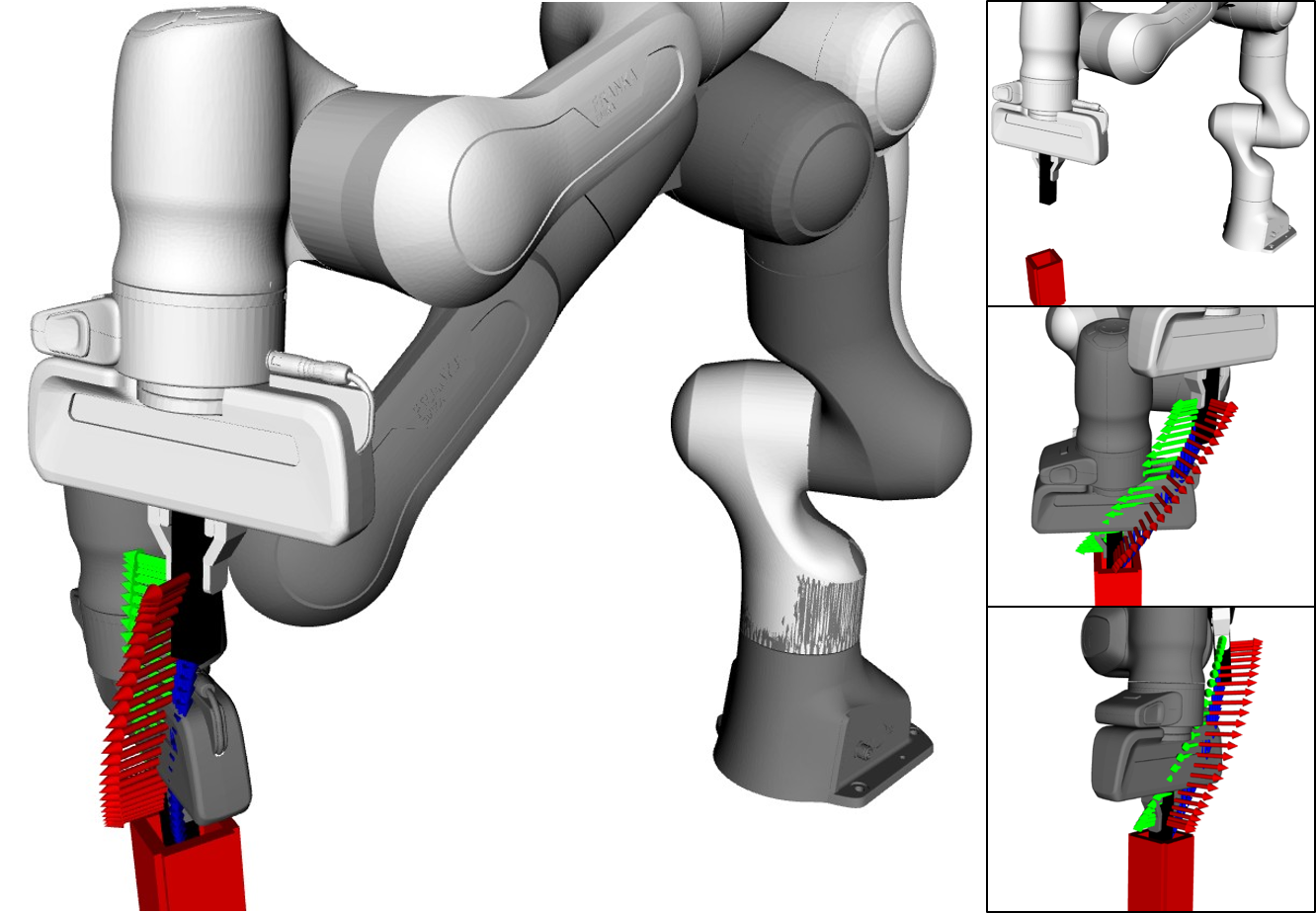} 
    \caption{Experiment setup.
    The task is to insert an object with a polytope geometric shape into a hole while maintaining a non-conservative approach to collision avoidance with the obstacles that form the hole.}
    \label{fig:cover_figure}
\end{figure}
The objective of reaching behavior is commonly represented as a point target, but this can be generalized to encompass a geometrically defined region or even an object. Traditional box constraints applied to system state and control variables also inherently involve geometric shapes, albeit in configuration space. 
While Euclidean projection~\cite{boyd2004convex} serves as an efficient technique for handling constraints in configuration space, its importance is often underestimated in the broader context of robotic behaviors. 
Indeed, the generation of these behaviors is closely and fundamentally associated with a geometric problem~\cite{low2023geometric}: it is predicated on both distance and its derivatives, which are intrinsically linked to the geometric shapes, sizes, and positions of robots and nearby objects. Projection techniques naturally align with this geometric framework, reinforcing their relevance.

Motivated by the challenges outlined, we introduce the concept of the geometric projector to capture geometric constraints and reformulate the optimization problem with the augmented Lagrangian method.
The key contributions are as follows:
\begin{enumerate}
    \item An efficient and user-friendly approach for designing geometric constraints, leveraging the geometric projectors.
    \item A unified, composable optimization framework that is applicable to a broad range of robots and behaviors.
    \item Our method is validated through numerical simulations and real-world experiments, with benchmarks and analyses that highlight its effectiveness.
\end{enumerate}

\section{Related work}
Considering the exact geometric shapes of robots and surrounding objects enables safer behavior without being overly conservative.
Instead of approximating the shapes by circles or ellipses,
one way to deal with polytopic shapes is to formulate a mixed-integer problem~\cite{grossmann2002review}.
However, the complexity and inefficiency limit its application.
Another popular branch is dual approaches~\cite{zhang2020optimization,bauschke2011convex} which handle it at the cost of additional dual variables.
Another branch techniques for collision avoidance are control barrier functions ~\cite{ames2019control}, which have also been applied to mobile robots~\cite{thirugnanam2023nonsmooth} and robot manipulators~\cite{dai2023safe} to generate safety-aware behaviors.
In our work, we show that this feature is easy to integrate and that we do not require smooth safety constraints.

In the context of shaping behaviors and composable framework,
geometric fabrics~\cite{van2022geometric} shape behaviors by bending geometry generators~\cite{ratliff2020optimization}.
Differentiable maps based on lower dimensional task variables can be designed separately and then pulled back to configuration space together and solved in a weighted average fashion.
Other energy-based methods~\cite{urain2021composable} also share the same similarities.
However, these methods typically do not consider horizons, while our work can be both reactive and predictive.
Projection~\cite{boyd2004convex} has been widely used in the collection of work mentioned above, as well as other robot applications or computer graphics~\cite{macklin2020local}.
In work~\cite{giftthaler2017projection}, M Giftthaler~\emph{et al.}~applied projection to 
handle equality constraints.
Based on the recent advance in augmented Lagrangian method for problems with geometric constraints~\cite{jia2023augmented},
H Girgin~\emph{et al.} applied it to point robot 
applications in~\cite{girgin2023projection}.
In contrast to their work, we
focus on a more general framework based on the concept of geometric projectors which consider more general projections between geometries with different shapes, sizes, positions and velocities capturing geometric constraints in robot applications.

\section{Problem formulation}\label{sec:problem formulation}
In this section, we establish the notions we will use in the GeoPro framework. 
We use $x$ to denote scalars, $\boldsymbol{x}$ for vectors, $\boldsymbol{X}$ for matrices, $\boldsymbol{\mathcal{X}}$ for tensors.
Subsequently, we introduce the concept of GeoPro and formalize it as a general nonlinear constrained optimization problem.

\subsection{Geometric Constrained Optimization}
We consider a general constrained nonlinear optimization problem of the form
\begin{subequations}\label{OCP}
    \begin{align}
    \min _{\boldsymbol{x}, \boldsymbol{u}}
            & \quad c(\boldsymbol{x}, \boldsymbol{u})=\sum_{k=0}^{N-1} l\left(\boldsymbol{x}_{k}, \boldsymbol{u}_{k}\right)  \nonumber \\
    \label{OCP.model}   \text { s.t. }  & \boldsymbol{x}_{k+1} = f(\boldsymbol{x}_k,\boldsymbol{u}_k),\\ 
    \label{OCP.constraitns} &g_i(\boldsymbol{x}_k)\in \mathcal{C}_i, \quad \forall i=1, \ldots, N_p, \\
    \label{OCP.xu} & \boldsymbol{x} \in \mathcal{D}_x, \quad \boldsymbol{u} \in \mathcal{D}_u
    \end{align}	
\end{subequations}%
where $N$ is the time horizon, $N_p$ is the number of constraints.
The cost function $l: \mathbb{R}^{n_x}\times \mathbb{R}^{n_u} \rightarrow \mathbb{R}$ depends on the robot tasks. The system dynamics $f:\mathbb{R}^{n_x}\times\mathbb{R}^{n_u}\rightarrow \mathbb{R}^{n_x}$ can be obtained by either Euler method or 4-th Runge–Kutta method. 
$g_i: \mathbb{R}^{n_x}\rightarrow \mathbb{R}^{n_i}$ is $i$-th continuously differentiable\footnote{It is assumed that only first-order derivative information is required for $g_i(\cdot)$} functions. 
The set $\mathcal{C}_i \in \mathbb{R}^{n_i}$ is nonempty, closed and normally convex.
The set $\mathcal{D}_x \in \mathbb{R}^{n_x}$ and $\mathcal{D}_u\in \mathbb{R}^{n_u}$ are only assumed to be nonempty and closed. 
The geometric nature of these constraints is captured by the sets defined in equations (\ref{OCP.constraitns})-(\ref{OCP.xu}).
The inequality and equality constraints can be described as $\mathcal{C}_i:=\mathbb{R}^{n_i}_{-}\times \{0\}$, which encodes $g_i(\boldsymbol{x})\leq 0$ and $g_i(\boldsymbol{x})=0$.
A Euclidean projection $P$ mapping a point onto a set is often given by $P_{\mathcal{C}_i}(\boldsymbol{x}) := \textstyle{\operatorname{argmin}_{\boldsymbol{z} \in \mathcal{C}_i}} \|\boldsymbol{z}-\boldsymbol{x}\|
$. 
The distance between a point $\boldsymbol{x}$ and a set $\mathcal{C}_i$ is denoted by $d_{\mathcal{C}_i}(\boldsymbol{x}):=||P_{\mathcal{C}_i} - \boldsymbol{x}||$.
Further, let $\textup{sd}_{\mathcal{C}_i}: \mathbb{R}^{n}\rightarrow \mathbb{R}$ be the signed distance function to $\mathcal{C}_i$:
\begin{equation}
\textup{sd}_{\mathcal{\mathcal{C}}_i}(\boldsymbol{x}):= \begin{cases}d_{\mathcal{C}_i}(\boldsymbol{x}), & x \in \mathbb{R}^n \backslash \mathcal{C}_i \\ -d_{\mathbb{R}^n \backslash \mathcal{C}_i}\left(\boldsymbol{x}\right), & x \in \mathcal{C}_i
\end{cases}
\end{equation}

Based on the aforementioned definitions, we introduce a generalized projection operator, denoted as $\mathcal{P}$ and referred to as the Geometric Projector (GeoPro), to map $g_i(x)$ onto $\mathcal{C}_i$ and encapsulate the geometric constraints.

\begin{remark}
    In~\cite{jia2023augmented}, the set $\mathcal{C}_i$ needs to be strictly convex.
    We found that for some non-convex sets, if we design projection of $g_i$ onto $\mathcal{C}_i$ to be unique, this condition can be relaxed to non-convex sets like the boundary of a circle.
\end{remark}

\subsection{Geometric Projector}
Let \( \Omega \subseteq \mathbb{R}^n \) denote the task space. 
Each geometry within this task space is defined by two elements: its state vector and its geometric shape. The state vector at time \( k \) is denoted as \( \boldsymbol{x}_k \) for robots, \( \boldsymbol{y}_k \) for obstacles, and \( \boldsymbol{p}_k \) for targets. Correspondingly, the geometric shape of each of these entities at time \( k \) occupies a specific subset of \( \Omega \), denoted by \( \mathcal{G}(\cdot) \).\footnote{The notation \( \mathcal{G} \) is used generically to refer to both obstacles and targets unless specified otherwise.}.
Our focus lies on three primary classes of objects: i): Robots, represented by \( \mathcal{G}_{R} \), ii): Obstacles, represented by \( \mathcal{G}_{O} \), iii): Targets, represented by \( \mathcal{G}_{T} \).
The composite geometries of $N_o$ obstacles and $N_t$ targets at time $k$ can be expressed as
\begin{equation}
\mathcal{G}_O\left(\boldsymbol{y}_k\right):=\bigcup_{i=1}^{N_o} \mathcal{G}_O^i \subset \Omega, \quad \mathcal{G}_T\left(\boldsymbol{p}_k\right):=\bigcup_{i=1}^{N_t} \mathcal{G}_T^i \subset \Omega,
\end{equation}
where \( \mathcal{G}_O^i \) and \( \mathcal{G}_T^i \) represent the individual geometries of the \(i^{th}\) obstacle and target, respectively.
Since we design behaviors for robots,
a robot-centered GeoPro, denoted as $\mathcal{P}_{\mathcal{G}}^{\mathcal{G}_R(\boldsymbol{x}_k)}$is formally defined as
\begin{equation}\label{GeoPro}
    \mathcal{P}: \mathcal{G}_R\left(\boldsymbol{x}_k\right) \times \mathcal{G} \rightarrow \mathcal{G}_R\left(\tilde{\boldsymbol{x}}_k\right),
\end{equation}
where $\tilde{\boldsymbol{x}}_k \in \mathbb{R}^n$ is the projected state vector of the robot at time $k$, and $\mathcal{G}$ allows to be $\varnothing$.
Through the design of GeoPro, various robot behaviors can be modeled. Specifically, we identify three main classes of behaviors: 
\begin{itemize}
    \item Safety behaviors, which aim to prevent collisions,
    \item Goal-reaching behaviors, that guide the robot toward a specific destination,
    \item Self-limiting behaviors, that restrict the robot's actions and states based on its capabilities.
\end{itemize}

\subsection{Robot Behaviors based on GeoPro}
Collision avoidance is a critical aspect of robotic behavior, ensuring the safety of both the robot and surrounding objects. 
For safety behaviors, denoted as $\mathcal{B}_{\textup{safe}}$, we formally define the following GeoPro $\mathcal{P}_{\mathcal{G}_{O}(\boldsymbol{y}_k)}^{\mathcal{G}_R(\boldsymbol{x}_k)}$:
\begin{align}
    &\mathcal{P}: \mathcal{G}_R\left(\boldsymbol{x}_k\right) \times \mathcal{G}_O\left(\boldsymbol{y}_k\right) \rightarrow \mathcal{G}_R\left(\tilde{\boldsymbol{x}}_k\right), \\
    & \mathcal{G}_{R}(\tilde{\boldsymbol{x}}_k) \cap \mathcal{G}_{O}(\boldsymbol{y}_{k}) = \varnothing,
\end{align}
which ensures that the geometries of the robot and the obstacles do not intersect.

In the context of goal-reaching behaviors, denoted by $\mathcal{B}_{\textup{reach}}$, we employ the following GeoPro $\mathcal{P}_{\mathcal{G}_{T}(\boldsymbol{p}_k)}^{\mathcal{G}_R(\boldsymbol{x}_k)}$: 
\begin{align}
    &\mathcal{P}: \mathcal{G}_R\left(\boldsymbol{x}_k\right) \times \mathcal{G}_T\left(\boldsymbol{p}_k\right) \rightarrow \mathcal{G}_R\left(\tilde{\boldsymbol{x}}_k\right), \\
    &\mathcal{G}_{R}(\boldsymbol{\tilde{x}}_k) \cap \mathcal{G}_{T}(\boldsymbol{p}_{k}) = \mathcal{G}_{\textup{reach}},
\end{align}
where $\mathcal{G}_{\textup{reach}}$ can vary from a simple geometric point to elaborate geometries, such as residing within a manifold dictated by $\mathcal{G}_{T}(\boldsymbol{y}_{k})$.

To control robot self-behaviors, denoted by $\mathcal{B}_{\textup{self}}$.
We suggest to use a Euclidean projection:
\begin{align}\label{eq: self}
    &P_{\mathcal{D}_x}(\boldsymbol{x}_k): \boldsymbol{x}_k \times \mathcal{D}_x \rightarrow \tilde{\boldsymbol{x}}_k\in \mathcal{D}_x, \\
    &P_{\mathcal{D}_u}(\boldsymbol{u}_k): \boldsymbol{u}_k \times \mathcal{D}_u \rightarrow \tilde{\boldsymbol{u}}_k\in \mathcal{D}_u.
\end{align}

Such projectors can be used to enforce constraints like speed or joint angle limits in robotics.

In summary, we can systematize and generalize a diverse array of robot behaviors, denoted collectively as $\mathcal{B}=\{\mathcal{B}_{\textup{safe}}, \mathcal{B}_{\textup{reach}},\mathcal{B}_{\textup{self}}\}$. 
These behaviors can further be customized by composing GeoPros like $\mathcal{B}:=\mathcal{B}_{\textup{safe}}\land \mathcal{B}_{\textup{reach}}$.
\section{Method}\label{sec: method}
In this section, we derive the details of GeoPro for general behaviors.
Next, we introduce how to integrate GeoPro into an optimization problem by the augmented Lagrangian method.

\subsection{GeoPro for Point Robot Geometry Behaviors}
For simplicity, we drop time notation $k$.
Let $\mathcal{G}_R(\boldsymbol{x})$ be a point geometry,
we consider two representations of obstacles and targets: hyperplanes, denoted by $\mathcal{H}$, with two half-spaces, denoted by $\mathcal{H}^{+}$ and $\mathcal{H}^{-}$, and the signed distance field, denoted by $\mathcal{S}$.
The space in $\Omega$ occupied by $i$-th obstacle can be represented by 
\begin{equation}
\scalebox{1}{$\mathcal{G}^i_{O}:=\{\boldsymbol{y}\in \mathbb{R}^n: \boldsymbol{A}\boldsymbol{y}\leq \boldsymbol{b}\}$},
\end{equation}
where $\boldsymbol{A}=\left[\boldsymbol{a}_1, \ldots, \boldsymbol{a}_{N^e_O}\right]^{\top}\in \mathbb{R}^{N^e_O\times n }$, $N^e_O$ is the number of edges and $\boldsymbol{b} \in \mathbb{R}^{N^e_O}$.
Each pair $(\boldsymbol{a}_j, b_j)$ constructs a half-space denoted by outward $\mathcal{H}^{+}(\boldsymbol{a}_j, b_j)$ aligned with normal direction and inward $\mathcal{H}^{-}(\boldsymbol{a}_j, b_j)$
where $\mathcal{H}^{-}(\boldsymbol{a}_j, b_j):=\{\boldsymbol{y}\in \mathbb{R}^n: \boldsymbol{a}_j^{\top} \boldsymbol{p} \leq b_j\}$.
The obstacle $\mathcal{G}^i_O$ can also be described as $\bigcap^{N^e_O}_j \mathcal{H}^{-}\left(\boldsymbol{a}_j, b_j\right)$.
For safe behaviors $\mathcal{B}_{\textup{safe}}$,
we introduce a condition $C_1$: 
\begin{equation}
    C_1:= \bigwedge_{j=1}^{N^{e}_{O}} \left(\boldsymbol{a}_j^{\top} \boldsymbol{x} \leq b_j\right),
\end{equation}
if $C_1$ evaluates to $\textit{true}$, this indicates a potential for collision and necessitates corrective action. 
In this case, we project $\boldsymbol{x}$ onto the closest $\mathcal{H}(\boldsymbol{a}_j, b_j)$ to ensure safety, as described by 
\begin{equation}\label{GeoPro:collision_avoidance}
    \mathcal{P}_{\mathcal{G}_{O}(\boldsymbol{y})}^{\mathcal{G}_R(\boldsymbol{x})}\rightarrow \mathcal{G}_{R}(\tilde{\boldsymbol{x}}) := \mathcal{G}_{R}\left(\boldsymbol{x}-\frac{\boldsymbol{a_j}\left(\boldsymbol{a_j}^{\top} \boldsymbol{x}-b_j\right)}{\|\boldsymbol{a_j}\|_2^2}\right),
\end{equation}
so that if $C_1$ is $\textit{false}$, we retain the original state $\boldsymbol{x}=\tilde{\boldsymbol{x}}$.
Since we are considering a point-like robot geometry $\mathcal{G}_R$, the GeoPro operation is equivalent to $P_{\mathcal{G}_O}(\boldsymbol{x})$ in this case.

For reaching behaviors $\mathcal{B}_{\textup{reach}}$, even though the goal  is often specified as a point, we consider a more general target geometry as
\begin{equation}
    \mathcal{G}^i_{T}:=\{\boldsymbol{p}\in \mathbb{R}^n: \boldsymbol{Z}\boldsymbol{p}\leq \boldsymbol{r}\},
\end{equation}
where $\boldsymbol{Z}=\left[\boldsymbol{z}_1, \ldots, \boldsymbol{z}_{N^e_T}\right]^{\top}\in \mathbb{R}^{N^e_T\times n }$, $N^e_T$ is the number of edges and $\boldsymbol{r} \in \mathbb{R}^{N^e_T}$.
For contact and reach-inside behaviors, it is insufficient to use the hyperplanes $\mathcal{H}(\boldsymbol{z}_{j}, r_{j})$ as they extend infinitely.
Let $\mathcal{E}=\left[e_1, \ldots, e_{N^e_T} \right]$ be the edge set, and  $(\boldsymbol{e}^a_{j}, \boldsymbol{e}^b_j)$ be the vertices of the edge $e_j$.
The auxiliary scalar is
\begin{equation}
    h=\frac{(\boldsymbol{x}-\boldsymbol{e}^a_{j})(\boldsymbol{e}^b_j-\boldsymbol{e}^a_{j})}{\left|\boldsymbol{e}^b_j-\boldsymbol{e}^a_{j}\right|^2},
\end{equation}
which is then clipped to limit projected points bounded by $h\leftarrow\max (0, \min (1, h))$.
The final GeoPro will be:
\begin{equation}\label{GeoPro:reach}
    \mathcal{P}^{\mathcal{G}_R(\boldsymbol{x})}_{\mathcal{G}_T(\boldsymbol{p})}\rightarrow \mathcal{G}_R{(\tilde{\boldsymbol{x}})}:=\mathcal{G}_{R}(
    (1-h)\boldsymbol{x} + h e^b_j),
\end{equation}
where $\mathcal{G}^i_T(\boldsymbol{p}):=\bigcap^{N^e_T}_j \mathcal{E}\left(\boldsymbol{e}^a_j, \boldsymbol{e}^b_j\right)$.
To facilitate the discussion on reaching behaviors, we introduce two conditions $C_2$ and $C_3$:
\begin{equation}
 C_2:=\bigvee\limits_{j=1}^{N^{e}_T} \left(\boldsymbol{z}_j^{\top} \boldsymbol{x} > r_j \right),
 C_3:=\bigwedge\limits_{j=1}^{N^{e}_{T}} \left(\boldsymbol{z}_j^{\top} \boldsymbol{x} \leq r_j\right),
\end{equation}
where $C_2$ pertains to contact behaviors.
If $C_2$ evaluates to $\textit{true}$, we do projection~(\ref{GeoPro:reach}). 
$C_3$ is associated with the reach-inside behaviors, if $C_3$ evaluates to $\textit{false}$, we apply GeoPro~(\ref{GeoPro:reach}). 

GeoPro $\mathcal{P}$ also supports the implicit SDF $\mathcal{S}$, either implicit functions or neural networks.\footnote{Interested readers can refer to various SDF examples in  \href{https://iquilezles.org/articles/}{link}, which can directly be incorporated into our framework.
For SDF tutorials, please refer to \href{https://rcfs.ch/doc/rcfs.pdf}{RCFS.} 
More applications can be found in \href{https://arxiv.org/pdf/2307.00533.pdf}{link}.
} 
For instance, We can represent an SDF using splines. 
The advantage of this approach is the availability of analytical and smooth gradient information.
Assume the order of the polynomials is $r$ and we have $c$ control points, the matrix form of the SDF is $\mathcal{S}:=\boldsymbol{t}^\trsp M\Phi$, $\boldsymbol{t}\in \mathbb{R}^{r}$ is the time vector, and $M\in \mathbb{R}^{r\times c}$ is the characteristic matrix and $\Phi\in \mathbb{R}^{c}$. 
The gradient $\nabla \mathcal{S}$ can be computed efficiently.

For behaviors $\mathcal{B}_{\textup{safe}}$, since the $\mathcal{G}_{R}(\boldsymbol{x})$ is a point geometry, the condition is the sign of $\textup{sd}_{S}(\boldsymbol{x})$, if it is negative, the GeoPro will be
\begin{equation}
   \mathcal{P}^{\mathcal{G}_R(\boldsymbol{x})}_{\mathcal{G}_{O}(\boldsymbol{y})} \rightarrow \mathcal{G}_{R}(\tilde{\boldsymbol{x}}):=\mathcal{G}_{R}(\boldsymbol{x} - \nabla \mathcal{S}(\boldsymbol{x})\mathcal{S}(\boldsymbol{x})).
\end{equation}

For reaching behaviors $\mathcal{B}_{\textup{reach}}$, we only need to change the condition as $\textup{sd}_{S}(\boldsymbol{x})>0 $.

\subsection{GeoPro for General Robot Geometry Behaviors}
Consider the robot has a geometric shape described by hyperplanes instead of a point, defined as 
$\mathcal{G}_R(\boldsymbol{x}):= \{\boldsymbol{x}\in\mathbb{R}^n: \boldsymbol{V}\boldsymbol{x} \leq \boldsymbol{d} \}$, where $\boldsymbol{V}= \left[\boldsymbol{v}_1, \ldots, \boldsymbol{v}_{N^e_R}\right]^{\top} \in \mathbb{R}^{N^e_R\times n}$, $N^e_R$ is the number of edges of robot geometry and $\boldsymbol{d} \in \mathbb{R}^{N^e_R}$.
GeoPro turns the projection between two geometries into an Euclidean projection by exploiting a Minkowski sum operation~\cite{ericson2004real}.
Assume we have two geometries $\mathcal{G}_A$ and $\mathcal{G}_B$, the Minkowski sum is defined as: 
\begin{equation}
    \mathcal{M}^{\mathcal{G}_A}_{\mathcal{G}_B}:=\mathcal{G}_A \oplus \mathcal{G}_B:=\{\boldsymbol{x}=\boldsymbol{x}_A - \boldsymbol{x}_B \mid \boldsymbol{x}_A\in \mathcal{G}_A, \boldsymbol{x}_B\in \mathcal{G}_B\},
\end{equation}
where $\mathcal{M}^{\mathcal{G}_A}_{\mathcal{G}_B}\subset  \Omega_{c}=\mathbb{R}^{n}$.
There is one important property of Minkowski sum: if $\mathcal{G}_A$ and $\mathcal{G}_B$ are intersecting, the origin of $\Omega_c$, denoted as $\boldsymbol{0}_{\Omega_c}$, lies inside the Minkowski sum $\mathcal{M}^{\mathcal{G}_A}_{\mathcal{G}_B}$, if they are not intersecting, Euclidean projection $P_{\mathcal{M}^{\mathcal{G}_A}_{\mathcal{G}_B}}(\boldsymbol{0}_{\Omega_c})$ will bring these two geometries in contact.

For behaviors $\mathcal{B}_{\textup{safe}}$ between $\mathcal{G}_R(\boldsymbol{x})$ and $\mathcal{G}_O(\boldsymbol{y})$, the condition is:
\begin{equation}
    C_{5}:=\boldsymbol{0}_{\Omega{c}} \in \mathcal{M}^{\mathcal{G}_R(\boldsymbol{x})}_{\mathcal{G}_O(\boldsymbol{y})},
\end{equation}
if $C_5$ holds $true$, the GeoPro $\mathcal{P}^{\mathcal{G}_{R}(\boldsymbol{x})}_{\mathcal{G}_{O}(\boldsymbol{y})}$ is defined as:
\begin{equation}
    \mathcal{P}^{\mathcal{G}_{R}(\boldsymbol{x})}_{\mathcal{G}_{O}(\boldsymbol{y})} \rightarrow \mathcal{G}_{R}(\tilde{\boldsymbol{x}}):=\mathcal{G}_{R}(P_{\mathcal{M}^{\mathcal{G}_R}_{\mathcal{G}_O}}(\boldsymbol{0}_{\Omega_c})),
\end{equation}
where $P_{\mathcal{M}^{\mathcal{G}_R}_{\mathcal{G}_O}}(\boldsymbol{0}_{\Omega_c})$ can be computed by~(\ref{GeoPro:collision_avoidance}).
For the behavior $\mathcal{B}_{\textup{reach}}$, the condition is modified as:
\begin{equation}
    C_{6}:=\boldsymbol{0}_{\Omega{c}} \notin \mathcal{M}^{\mathcal{G}_R(\boldsymbol{x})}_{\mathcal{G}_T(\boldsymbol{p})},
\end{equation}
if $C_6$ is $true$, the GeoPro for reach behaviors is defined as:
\begin{equation}
    \mathcal{P}^{\mathcal{G}_{R}(\boldsymbol{x})}_{\mathcal{G}_{T}(\boldsymbol{p})} \rightarrow \mathcal{G}_{R}(\tilde{\boldsymbol{x}}):=\mathcal{G}_{R}(P_{\mathcal{M}^{\mathcal{G}_R}_{\mathcal{G}_T}}(\boldsymbol{0}_{\Omega_c})),
\end{equation}

In this section, we introduce the GeoPro as a foundational component for reformulating the constrained optimization problem~\ref{OCP}. To address projection-based constraints, we employ the augmented Lagrangian method, as detailed in~\cite{jia2023augmented}. To solve the resulting constrained subproblem, we utilize spectral projected gradient descent. This integrated algorithm is referred to as GeoPro-based ALSPG. Comprehensive details, including mathematical derivations and algorithmic steps, are provided in the Appendix

\section{experiments}
We evaluate the effectiveness and performance of the geometric projector framework in simulation, as well as in experiments with a 7-axis Franka Emika robot arm. 

\begin{figure}[t!]
  \centering
  \subfigure[$\mathcal{B}_{\textup{safe}}$]{\includegraphics[width=0.24\textwidth]{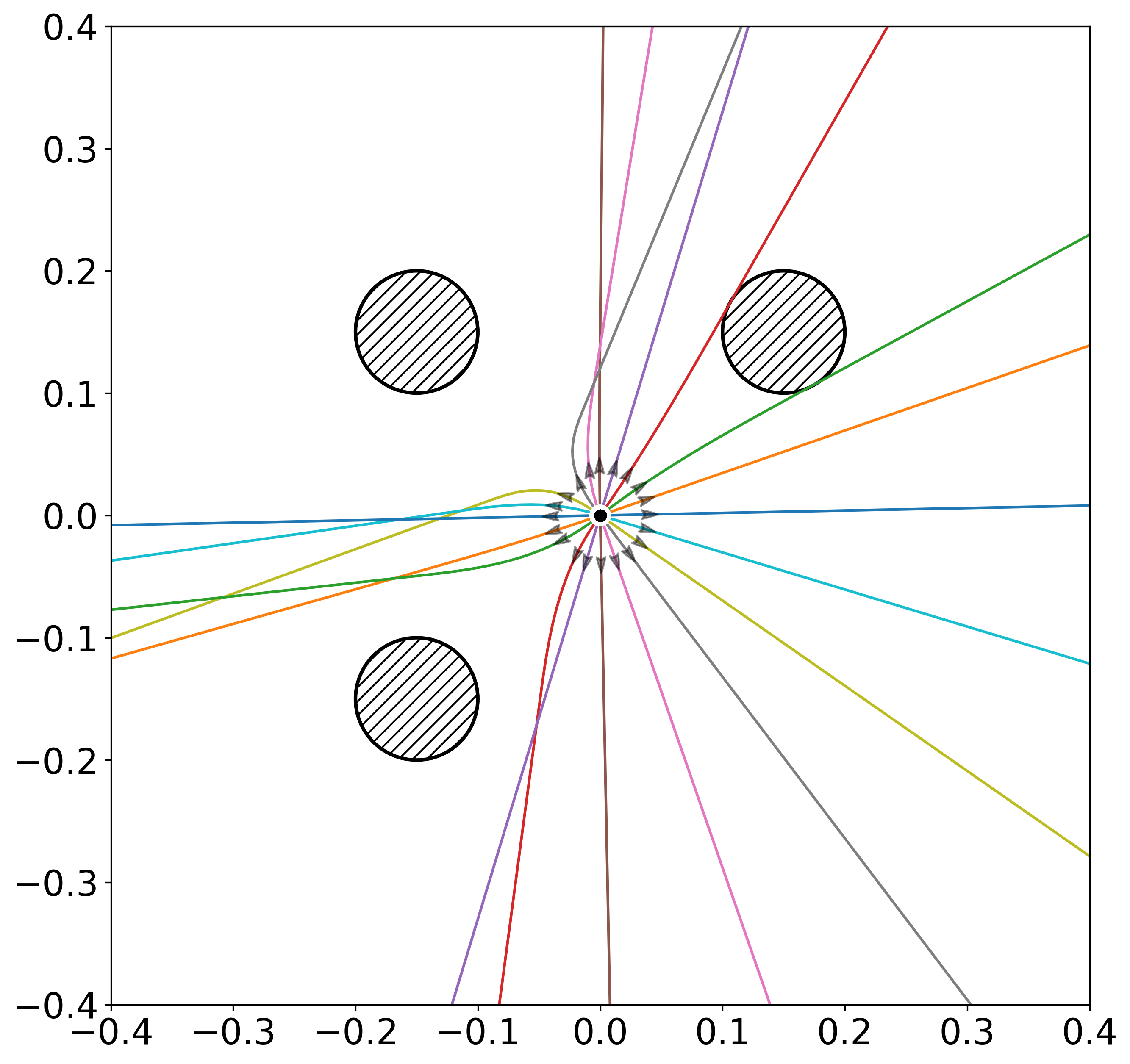}\label{fig:shape_01}}%
  \subfigure[$\mathcal{B}_{\textup{safe}}\land \mathcal{B}_{\textup{reach}}$]{\includegraphics[width=0.24\textwidth]{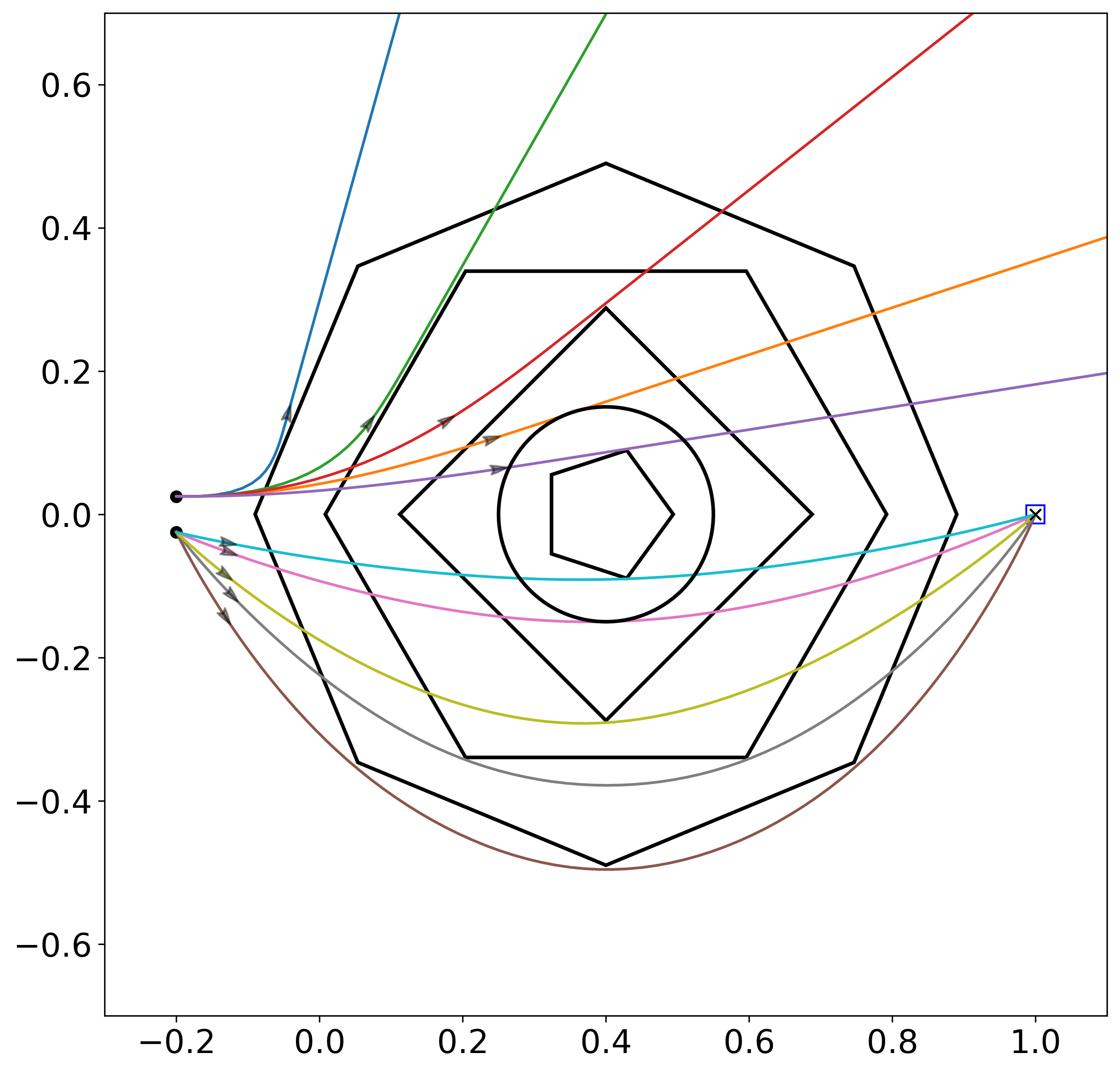}\label{fig:shape_02}}%
  \hfill
  \\
  \subfigure[$\mathcal{B}_{\textup{safe}}\land \mathcal{B}_{\textup{reach}} \land \mathcal{B}_{\textup{limit}}$]{\includegraphics[width=0.24\textwidth]{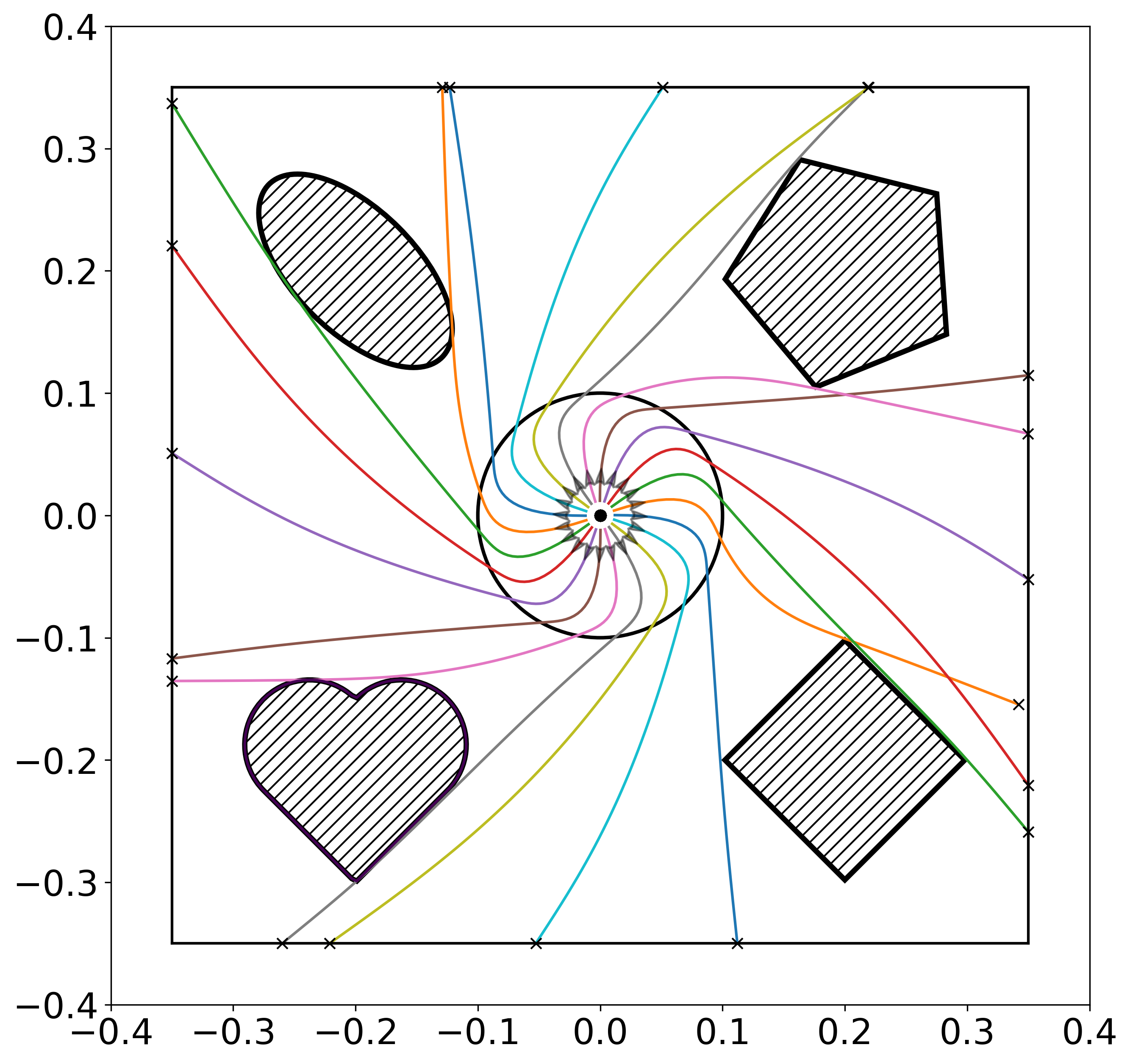}\label{fig:shape_03}}%
  \hfill
  \subfigure[$\mathcal{B}_{\textup{safe}}\land \mathcal{B}_{\textup{reach}}\land\mathcal{B}_{\textup{limit}}$]{\includegraphics[width=0.24\textwidth]{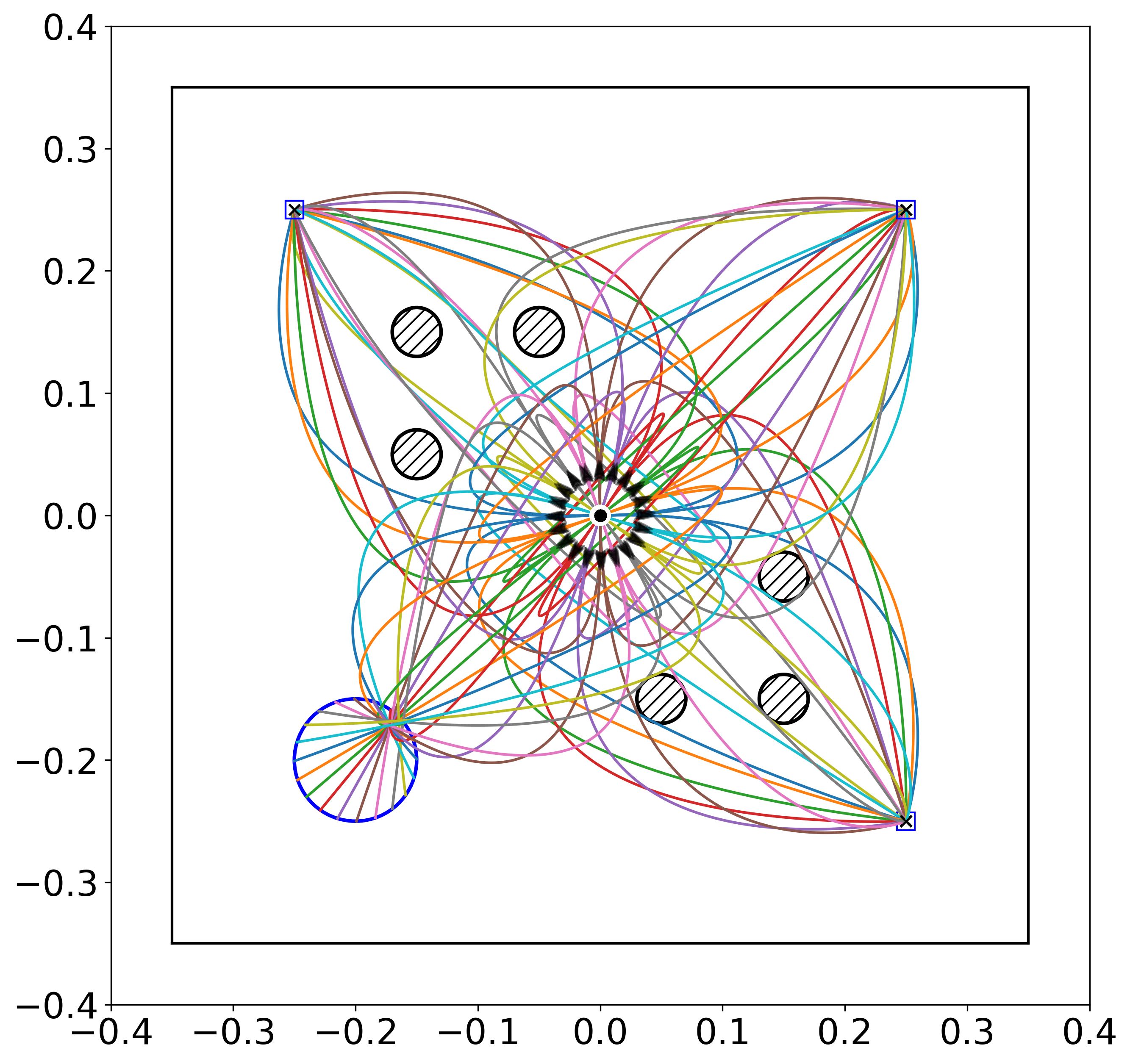}\label{fig:shape_04}}%
  \caption{Shape the robot behaviors by geometric projectors.
  (a) Demonstrates varying safety buffers in GeoPro to achieve conservative safe maneuvers.
  (b) Illustrates agents circumventing polytopic obstacles to consistently reach their goals.
  Polytope obstacles are very common in robotics.
 (c) Introduces subgoals and operational constraints for the robots; the robots are required to reach designated subgoals along the circle while adhering to either task-space or joint-space box limits to regulate velocities and accelerations.
 (d) Explores the capability of robots to form various goal shapes, useful in platoon planning and control.
 GeoPro further offers barrier-function-like safety mechanisms: as robots approach obstacles, their behavior becomes more conservative, while they proceed directly towards the goal when obstacles are distant.
 GeoPro's layered computational architecture allows for the integration of multiple objectives and constraints, thereby enabling diverse robot behaviors.
  For more details of shaping behaviors, please refer to \href{https://www.xueminchi.com/publications/geopro}{https://www.xueminchi.com/publications/geopro}.
  }
  \label{fig:shape behaviors}
\end{figure}

\subsection{Shape Robot Behaviors by GeoPro}
We begin by showing that the robot behaviors can be shaped by composing behaviors $\mathcal{B}:=\mathcal{B}_{\textup{safe}}\land \mathcal{B}_{\textup{reach}}\land \mathcal{B}_{\textup{limit}}$.
The system is a 2D point mass: 
\begin{equation}\label{eq:DI_model}
    \dot{c}_x=v_x, \hspace{0.4em} \dot{c}_y=v_y,\hspace{0.4em} \dot{v}_x=a_x, \hspace{0.4em} \dot{v}_y=a_y,
\end{equation}
where the system states are $\boldsymbol{x}=(c_x,c_y,v_x,v_y)$ with $\boldsymbol{u}=(a_x,a_y)$ as inputs.
In Fig.~\ref{fig:shape_01}, a set of 20 agents with initial conditions $(c_x, c_y)=(0,0)$ and $\|v\|=1$ pointing outward and angles are evenly spaced from $0$ to $2\pi$. 
Though the obstacle shapes are the same, we can design different GeoPro to have different safe behaviors.
The right upper obstacle is without any buffer while the safety buffer for the left corner and upper is 0.1 and 0.05 respectively.
In Fig.~\ref{fig:shape_02}, 5 agents on each side with the initial conditions $\boldsymbol{x}_{\textup{init}}=\left[-0.2, \pm 0.025, 0, 0\right]$ respectively, the goal for bottom 5 agents is $\boldsymbol{p}_{\textup{reach}}=\left[1, 0, 0, 0\right]$.
It is shown that when the goal-reaching behavior is considered, the bottom 5 agents converge to the goal while the upper 5 agents move without goals.
In Fig.~\ref{fig:shape_03}, a set of 20 agents have subgoals distributed in the circle.
The goal-reaching behavior is defined implicitly by asking the agent to touch the goal instead of specifying a time sequence explicitly.
The heart-shaped obstacle shows our framework is able to handle a variety of object geometries.
In Fig.~\ref{fig:shape_04}, a swarm of 80 agents is depicted, all of which share the same information about obstacles. For agents located in the right-upper corner, the safety measures differ depending on their direction of approach to the obstacles. When converging from the bottom side, the agents are permitted to approach the obstacles more closely. In contrast, when they are converging from the upper side, the agents employ behaviors consistent with control barrier functions to enhance safety.
In order to incorporate safety behaviors, we extend GeoPro by incorporating a safety function, denoted as $\psi(d(\boldsymbol{x}, \boldsymbol{u}))$.
The modified GeoPro is defined as: $\mathcal{P}^{\mathcal{G}_{R}}_{\mathcal{G}_{O}}\rightarrow\mathcal{G}_{R}(\Tilde{\boldsymbol{x}}+\psi(d(\boldsymbol{x}, \boldsymbol{u})))$.
Here, $\psi(d(\boldsymbol{x}, \boldsymbol{u}))$ is specified as:
\begin{equation}
\psi(d)= \begin{cases}0 & \dot{d}(\boldsymbol{x}, \boldsymbol{u}) \geq 0 \\ \gamma (d(\boldsymbol{x}, \boldsymbol{u})) & \dot{d}(\boldsymbol{x}, \boldsymbol{u})<0\end{cases}
\end{equation}
in this formulation, $d$ represents the distance metric, and $\dot{d}(\boldsymbol{x}, \boldsymbol{u})$ signifies its time derivative.
When the robot is approaching an obstacle $(\dot{d}(\boldsymbol{x}, \boldsymbol{u}))$, the function $\gamma(d(\boldsymbol{x}, \boldsymbol{u})$ is invoked to modify the state $\boldsymbol{x}$, effectively enhancing the safety behavior of the robot.
The $\gamma$ function can be a scalar or an extended class $\mathcal{K}_{\infty}$ function, offering flexibility in the choice of safety margins.
For the left-corner goal-reaching behavior, 20 agents are required to safely reach the circle-shaped goal.

\subsection{Non-holonomic Mobile Robots}
We demonstrate our framework on non-holonomic mobile robots with different geometric shapes without resorting to approximations.
For non-holonomic robots, we employ the following model:
\begin{equation}
    \dot{c}_x=v \cos (\theta), \quad \dot{c}_x=v \sin (\theta), \quad \dot{\theta}=w
\end{equation}
where system states are $\boldsymbol{x}=(c_x, c_y, \theta)$ with $\boldsymbol{u}=(v,w)$ velocity and turning rate as inputs.
The system behaviors are $\mathcal{B}:={\mathcal{B}_{\textup{safe}} \land \mathcal{B}_{\textup{reach}}}$. 
\begin{figure}[t!]
  \centering
  \subfigure[Rectangle]{\includegraphics[width=0.24\textwidth]{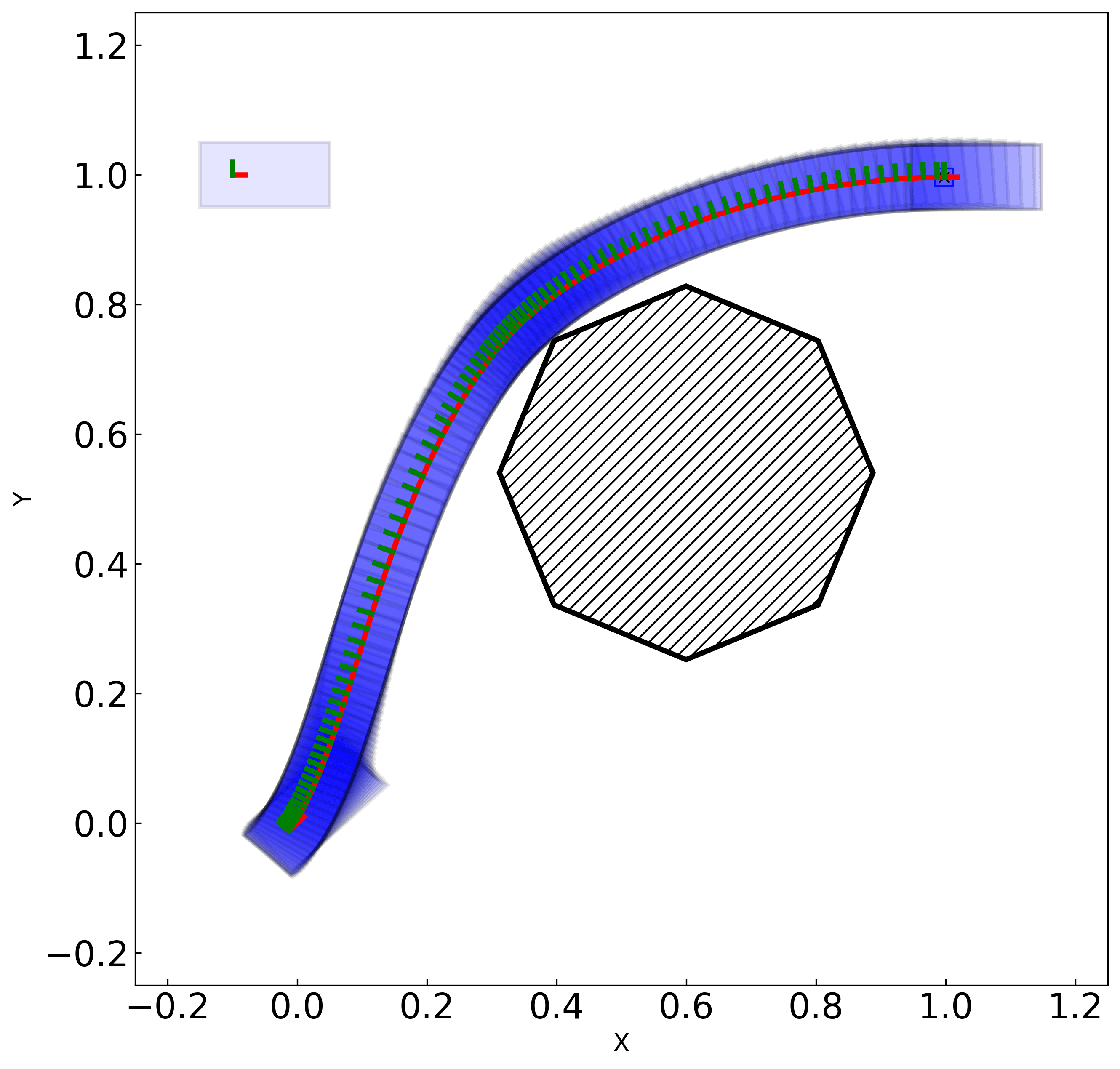}\label{fig:Du_rec}}%
  \hfill
  \subfigure[quadrilateral]{\includegraphics[width=0.24\textwidth]{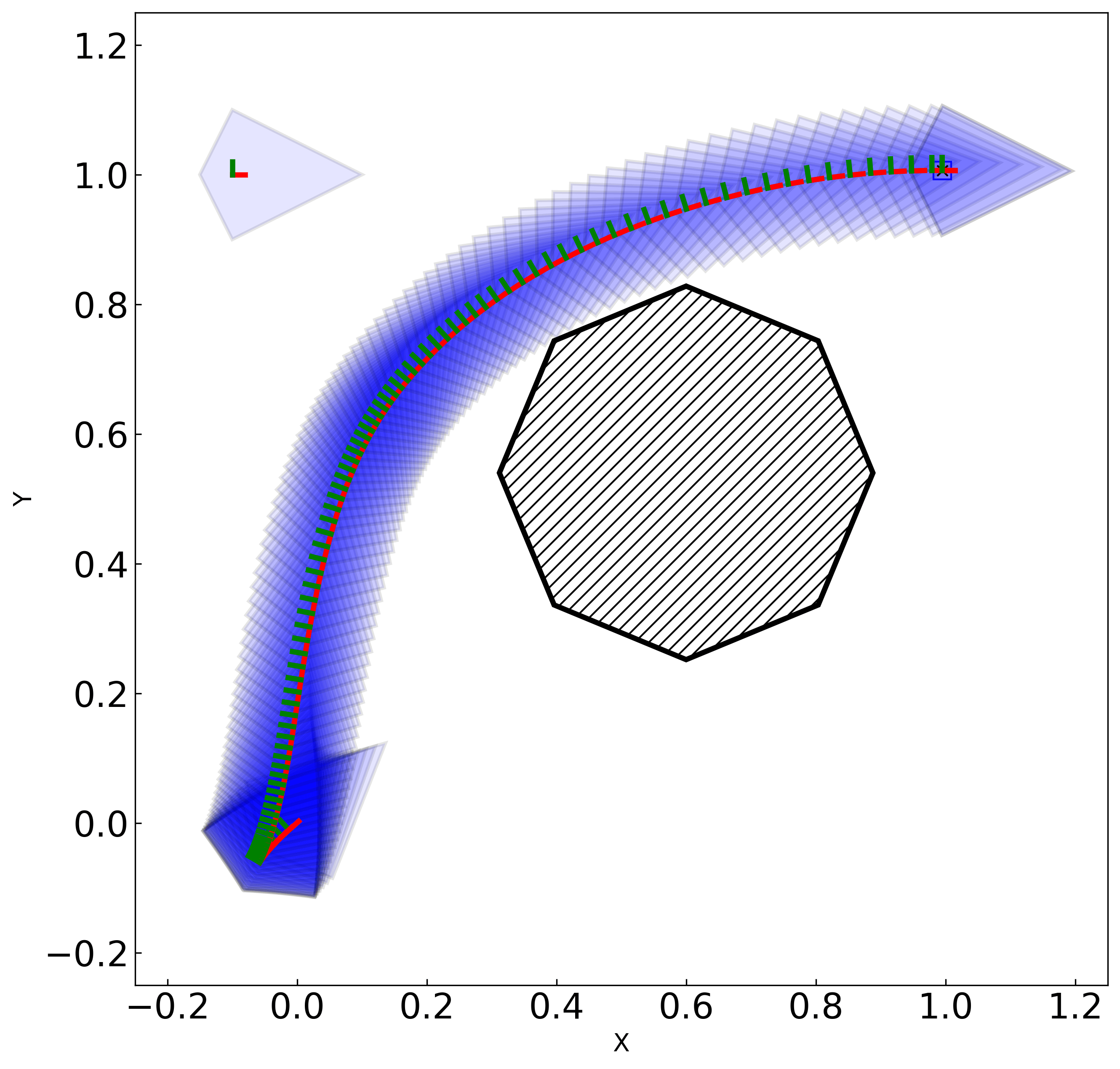}\label{fig:Du_quad}}%
  \\
  \subfigure[L shape]{\includegraphics[width=0.24\textwidth]{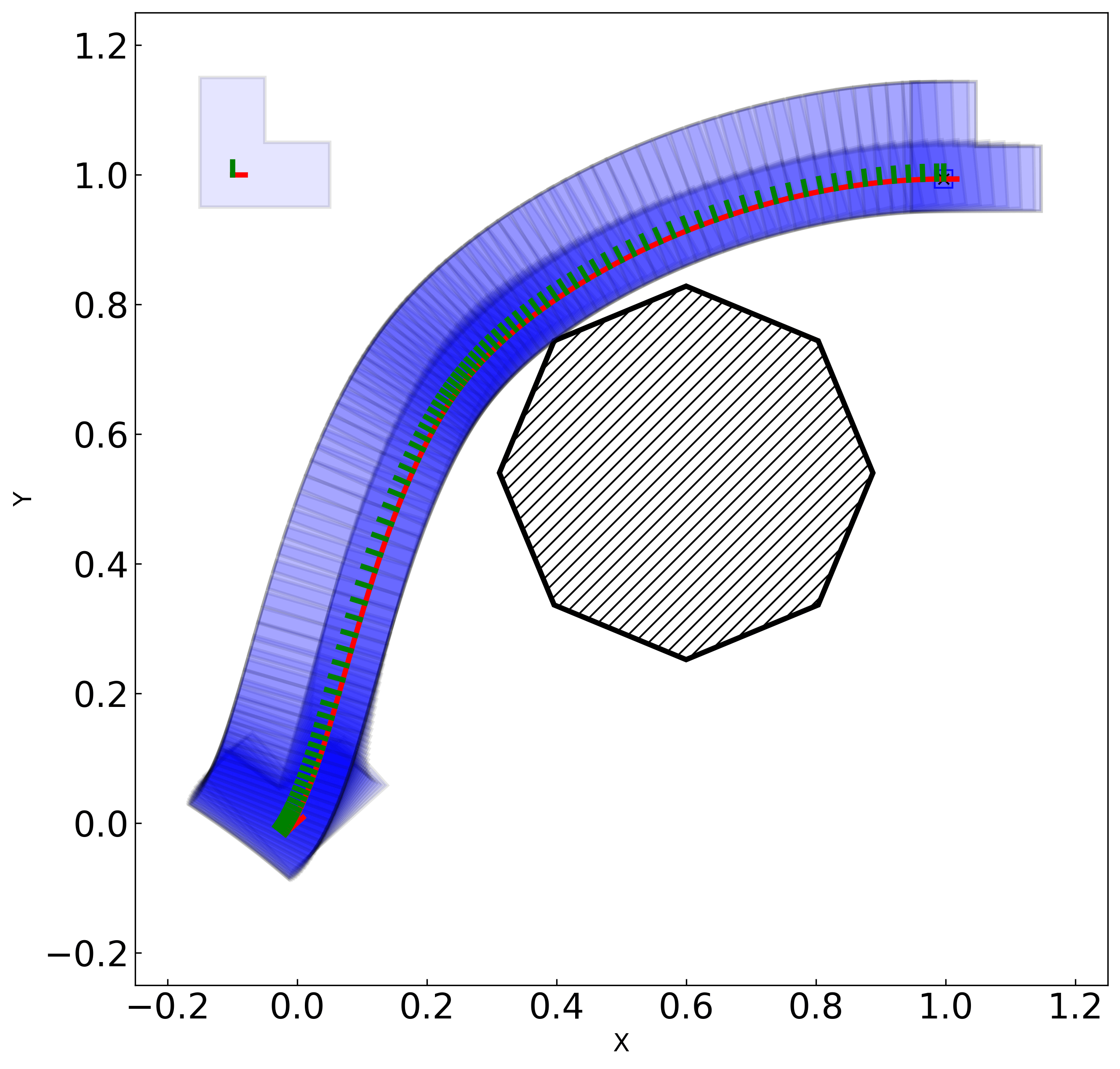}\label{fig:Du_L}}%
  \hfill
  \subfigure[Triangle]{\includegraphics[width=0.24\textwidth]{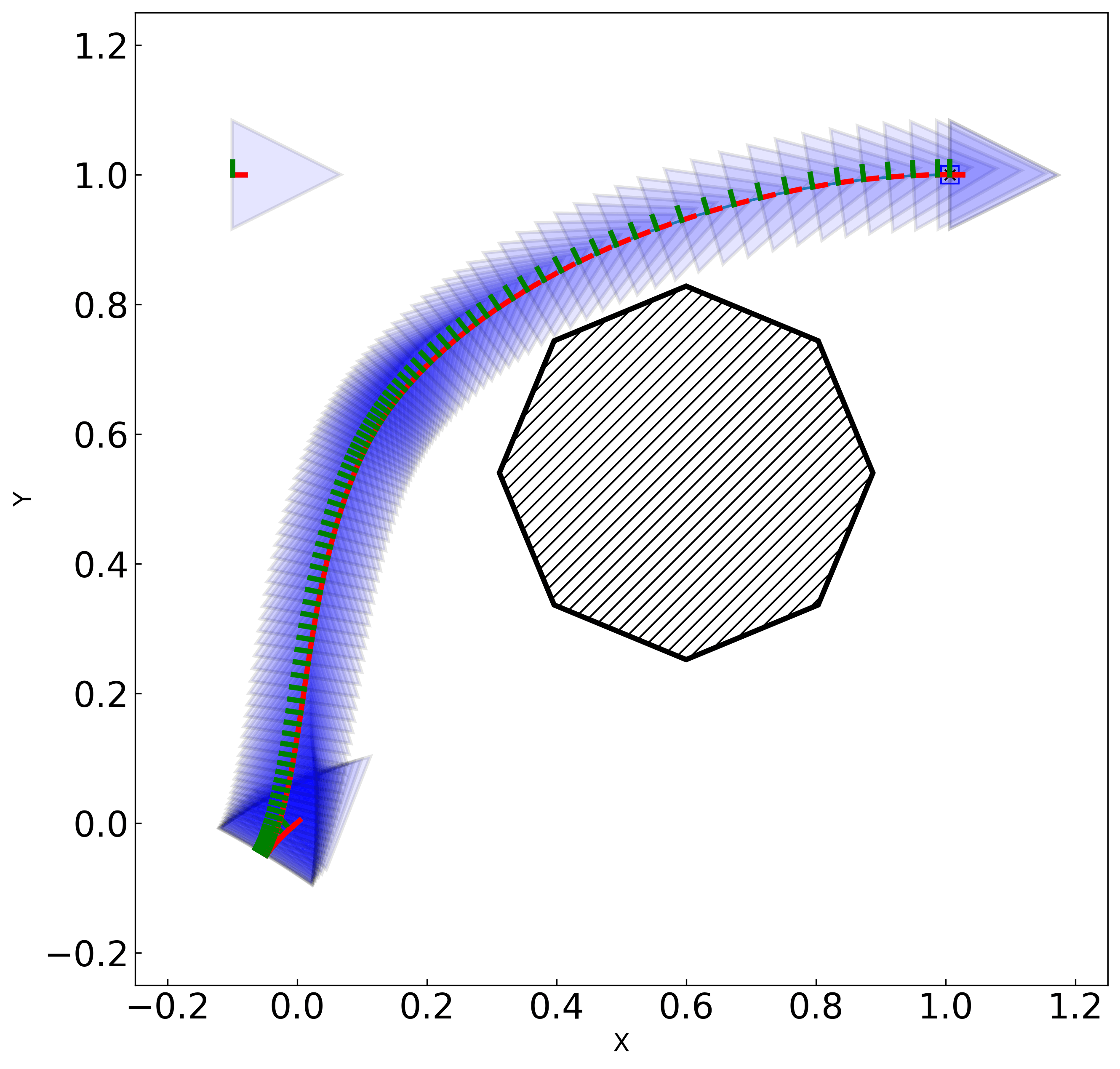}\label{fig:Du_triangle}}%
  \caption{Non-holonomic mobile robots with different geometric shapes.
  }
  \label{fig:Non-holonomic_robots}
\end{figure}
One key advantage of the GeoPro framework lies in its ability to handle collision avoidance without needing to formulate it as a twice-differentiable constraint\footnote{This is typically considered a requirement for smoothness in many problem formulations}.
Additionally, our approach does not require decomposing non-convex shapes like the L-shape into convex components, making the framework both flexible and versatile.
However, it should be noted that since the problem is inherently non-convex and we do not utilize any initial guesses, the optimality of the resulting trajectory cannot be guaranteed.

\subsection{Planar Arm}
\begin{figure}[t!]
  \centering
  \subfigure[$\mathcal{B}_{\textup{limit}}\land \mathcal{B}_\textup{reach}$]{\includegraphics[width=0.24\textwidth]{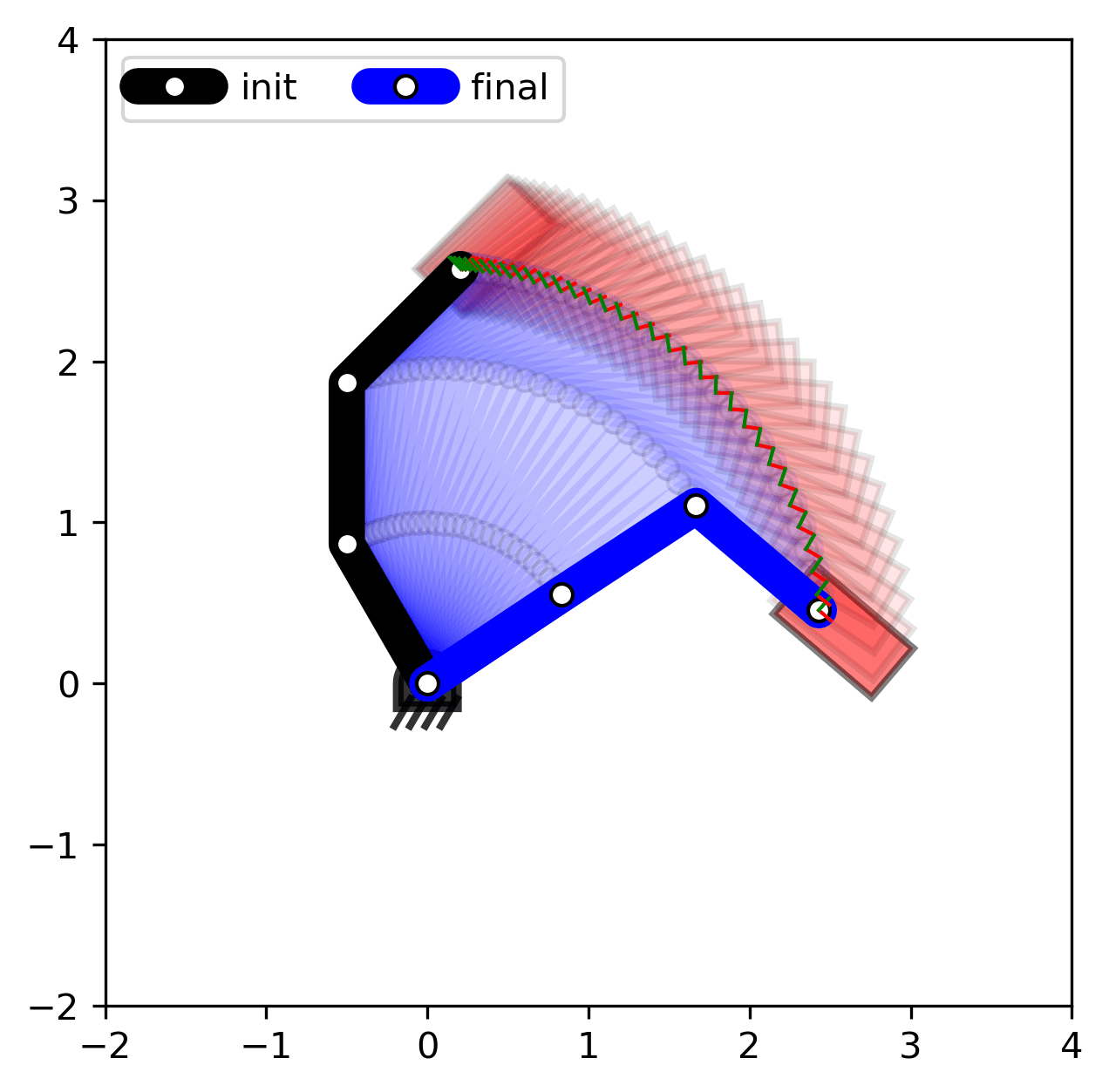}\label{fig:Ma_01}}%
  \hfill
  \subfigure[Straight line]{\includegraphics[width=0.24\textwidth]{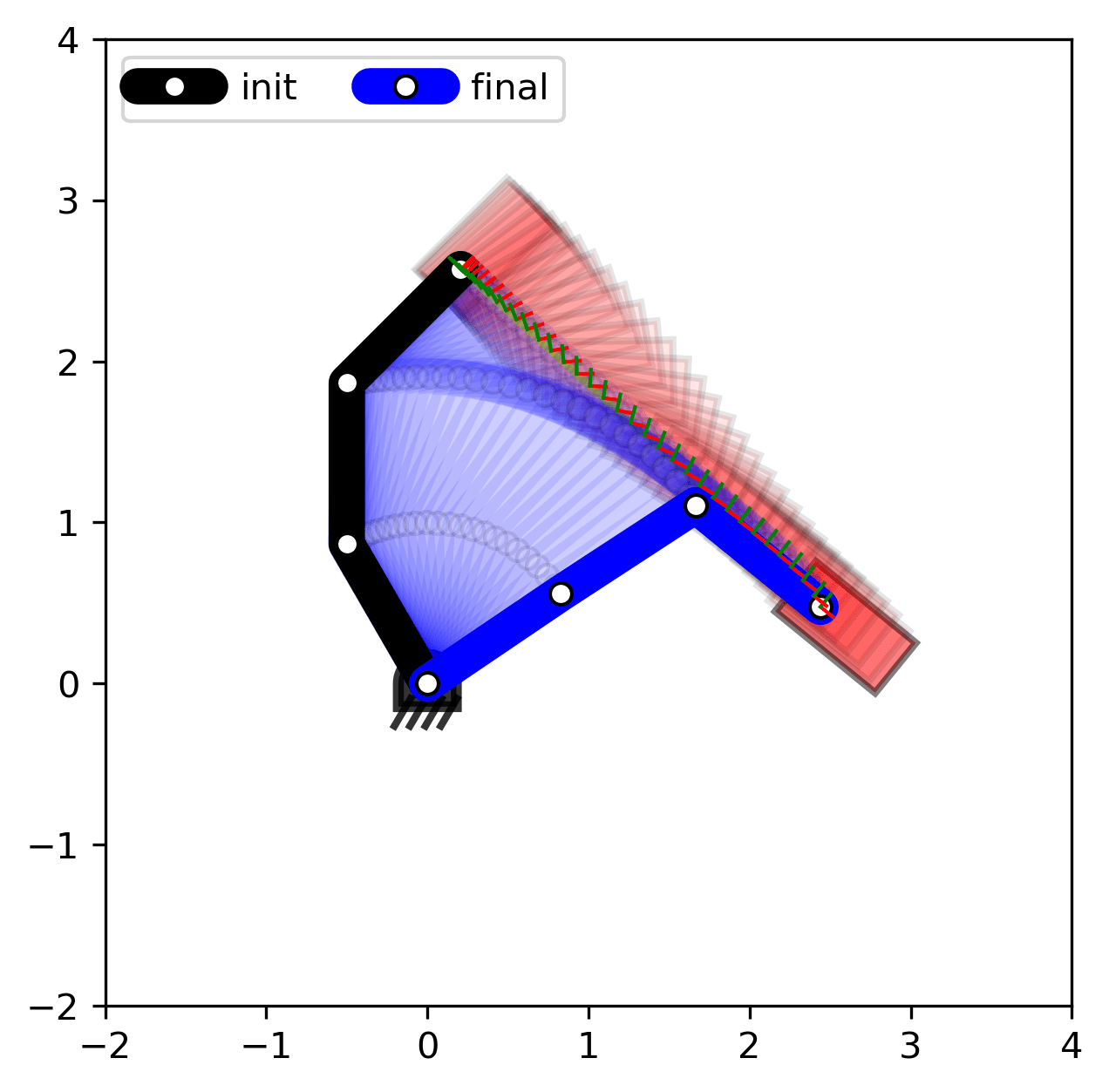}\label{fig:Ma_02}}%
  \\
  \subfigure[Surface]{\includegraphics[width=0.24\textwidth]{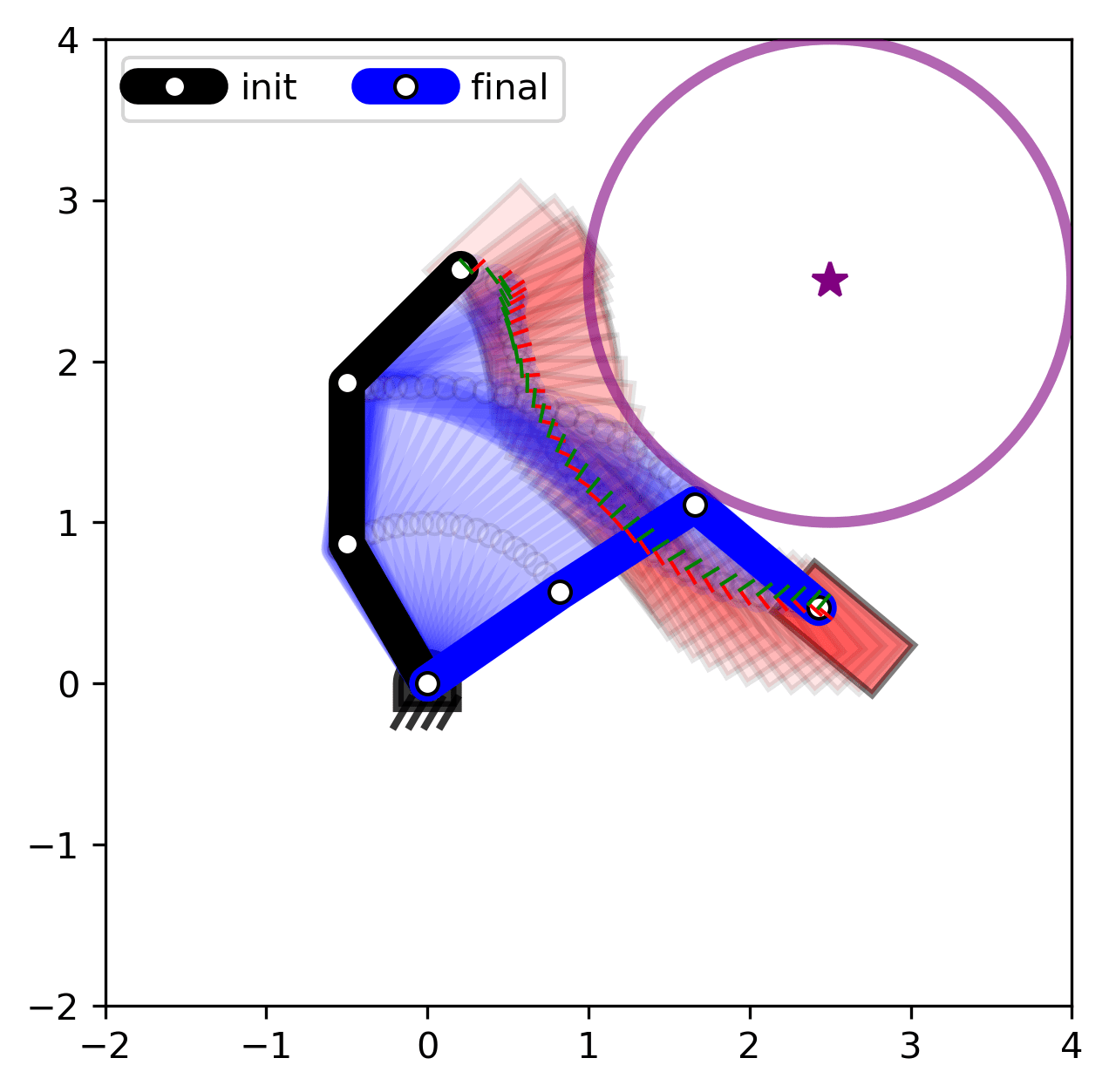}\label{fig:Ma_03}}%
  \hfill
  \subfigure[Region]{\includegraphics[width=0.24\textwidth]{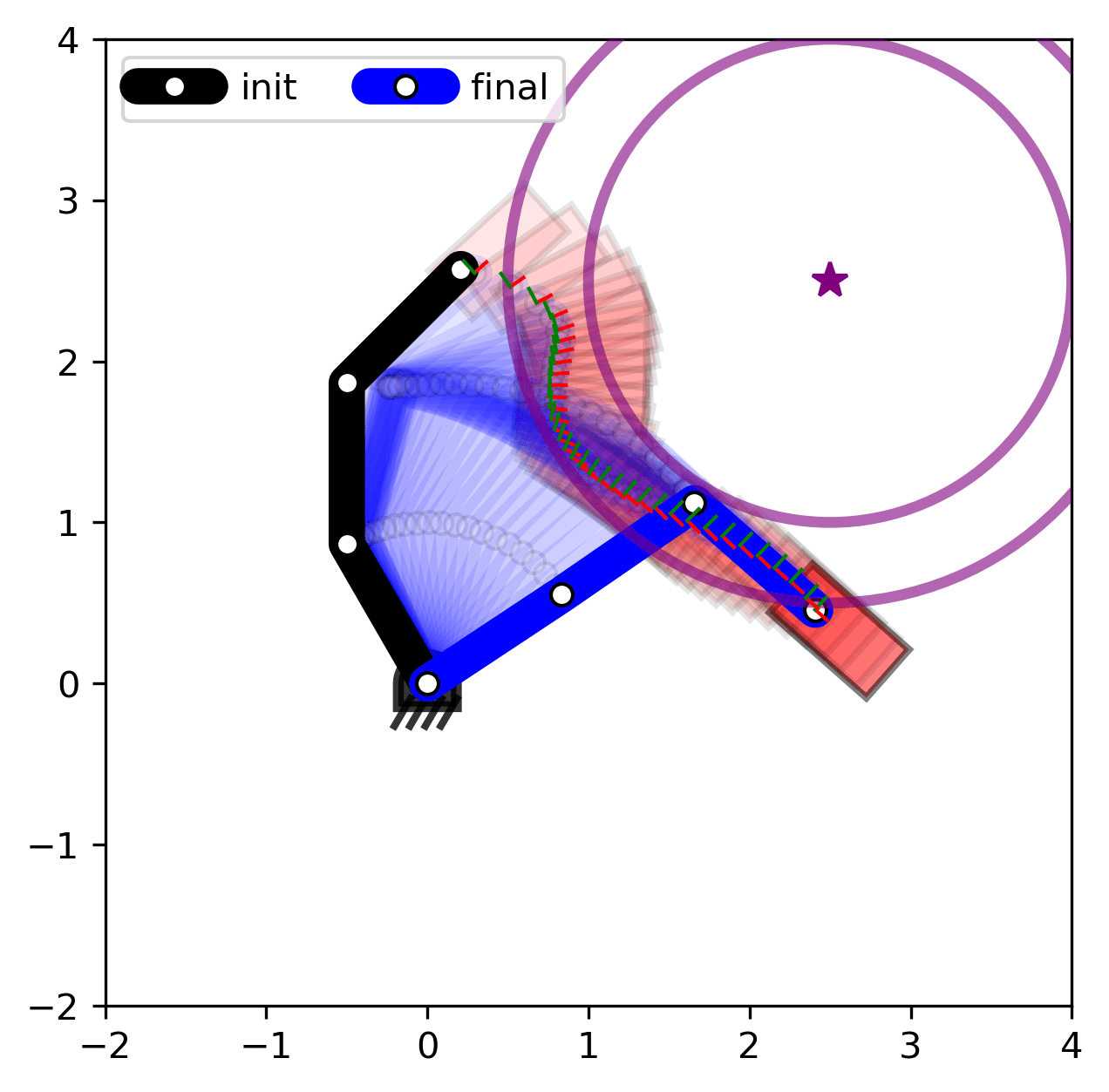}\label{fig:Ma_03_region}}%
  \\
  \subfigure[Insertion]{\includegraphics[width=0.24\textwidth]{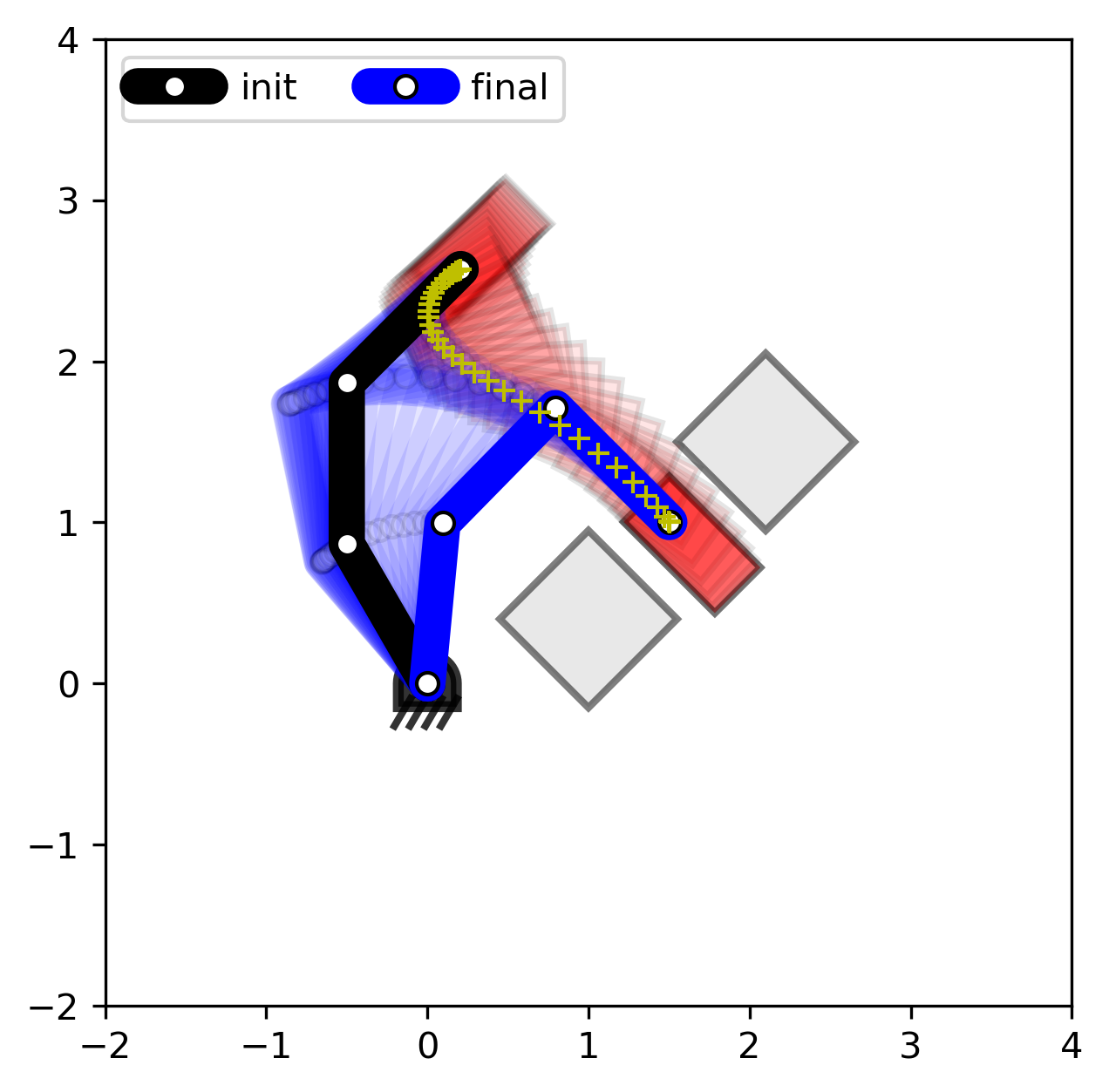}\label{fig:Ma_peg}}%
  \hfill
  \subfigure[Object centered]{\includegraphics[width=0.24\textwidth]{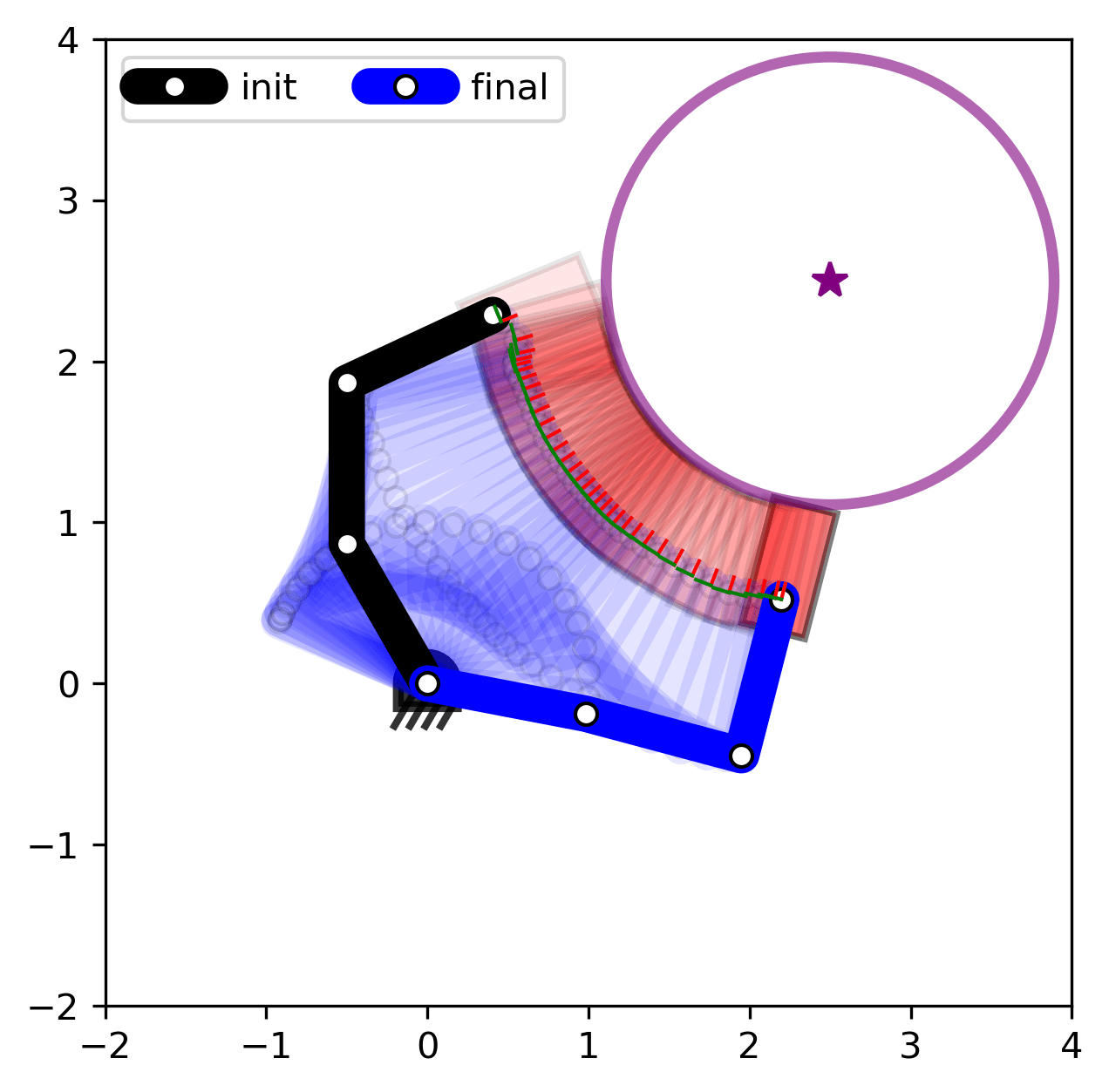}\label{fig:Ma_tele}}%
  \caption{GeoPro based planar arm behaviors.
  The rectangle geometry is fixed to the end-effector.
  }
  \label{fig:Planar_Arm}
\end{figure}

GeoPro can also be extended to rigid robots with kinematic chains to obtain specified behaviors, as illustrated in Fig.~\ref{fig:Planar_Arm}.
To demonstrate this capability, we focus on a 3-DOF planar arm.
The system model is
$\boldsymbol{x}=f(\boldsymbol{q})$\footnote{\href{https://robotics-codes-from-scratch.github.io/}{For more details and codes, refer to the Robotics Codes from Scratch toolbox.}}, with state vector as $\boldsymbol{x}=\left[q_1, q_2, q_3, \dot{q}_1, \dot{q}_2,\dot{q}_3, c_x, c_y, \theta \right]$ while the control input vector is $\boldsymbol{u}=\left[\ddot{q}_1, \ddot{q}_2, \ddot{q}_3 \right]$.
In Fig.~\ref{fig:Ma_01}, the $\mathcal{B}_{\textup{limit}}$ in joint space is $\boldsymbol{\ddot{q}}_{\textup{min}}\leq \boldsymbol{\ddot{q}}\leq\boldsymbol{\ddot{q}}_{\textup{max}}$.
Geometrically, the feasible region in joint space resembles a box centered at the origin $\boldsymbol{0}$.
GeoPro forces joint limits by projecting onto the feasible region.
The $\mathcal{B}_{\textup{reach}}$ behavior aims to reach a point with specified orientation, $\left[c_x, c_y, \theta\right]$ by projection.
The cost function for these behaviors only considers minimum efforts $\boldsymbol{u}$ with zero input as an initial guess.

In Fig.~\ref{fig:Ma_02}, an additional GeoPro requires the end-effector to reach the goal while following a specified line. 
Fig.~\ref{fig:Ma_03} shows a task where the end-effector maintains a specific distance from an object while also maintaining a certain pose upon reaching a designated point. Similarly, Fig.~\ref{fig:Ma_03_region} shows the end-effector being permitted to move within one manifold and outside another, while maintaining a specific orientation upon reaching a point.

Insertion tasks~\cite{shetty2021ergodic} are common in practice.
Our GeoPro-based framework can directly accommodate the geometric considerations inherent in peg-in-hole tasks, as shown in Fig.~\ref{fig:Ma_peg}.
This behavior combines $\mathcal{B}_{\textup{safe}}\land \mathcal{B}_\textup{reach}\land \mathcal{B}_\textup{limit}$, where $\mathcal{B}_{\textup{safe}}$ ensures that the peg attached to the end-effector avoids obstacles.
Object-centered tasks, which involve intricate geometric relationships between the robot and the object, can also be tackled. 
As illustrated in Fig.~\ref{fig:Ma_tele}, we define a behavior that allows the robot to move around a target while ensuring that the end-effector remains oriented towards that target.

\subsection{Autonomous Driving Benchmark}
\begin{figure}[t!]
  \centering
  \subfigure[Vertical Parking]{\includegraphics[width=0.23\textwidth]{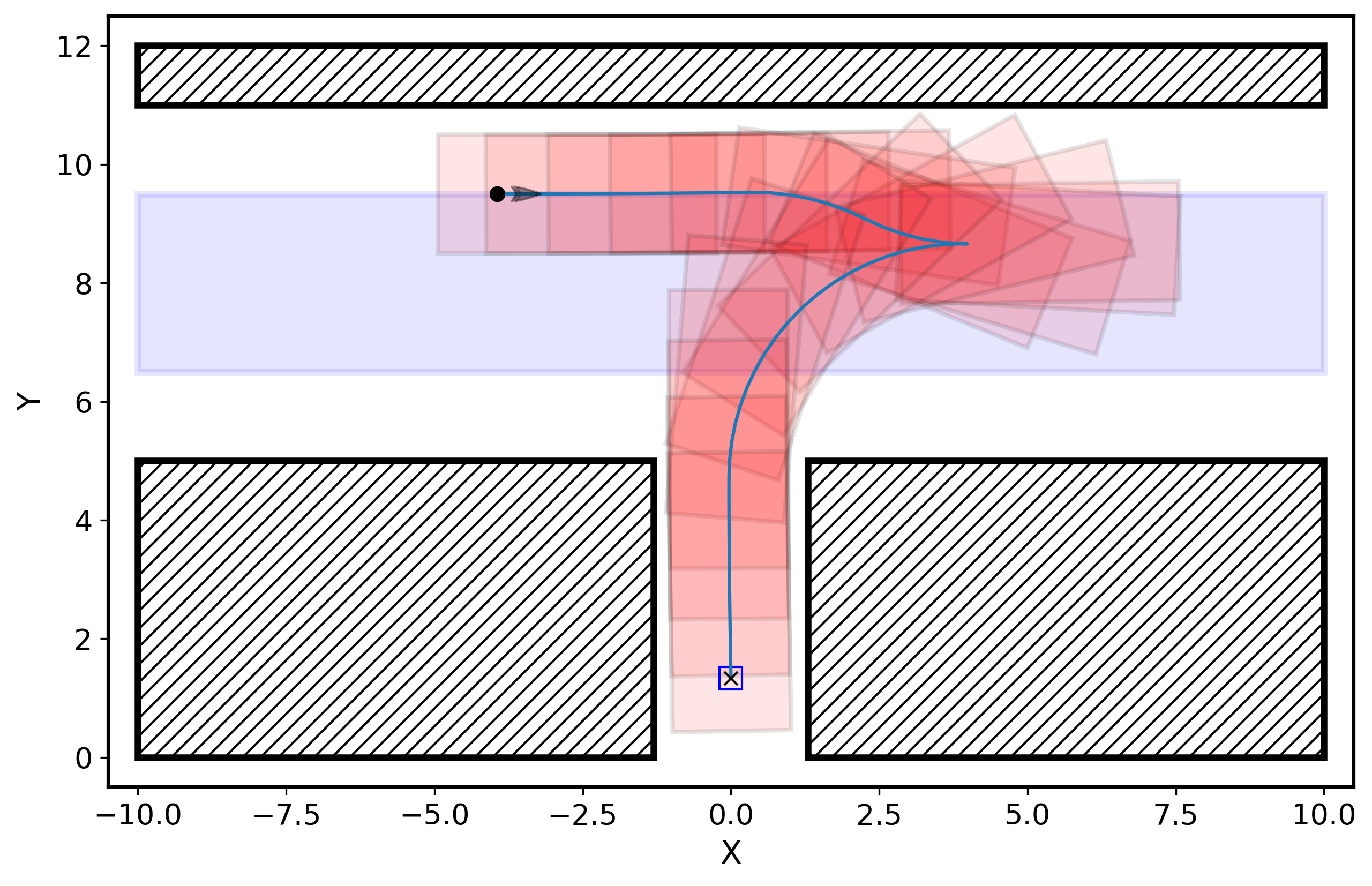}\label{fig:Vertical}}%
  \hfill
  \subfigure[Parallel Parking]{\includegraphics[width=0.23\textwidth]{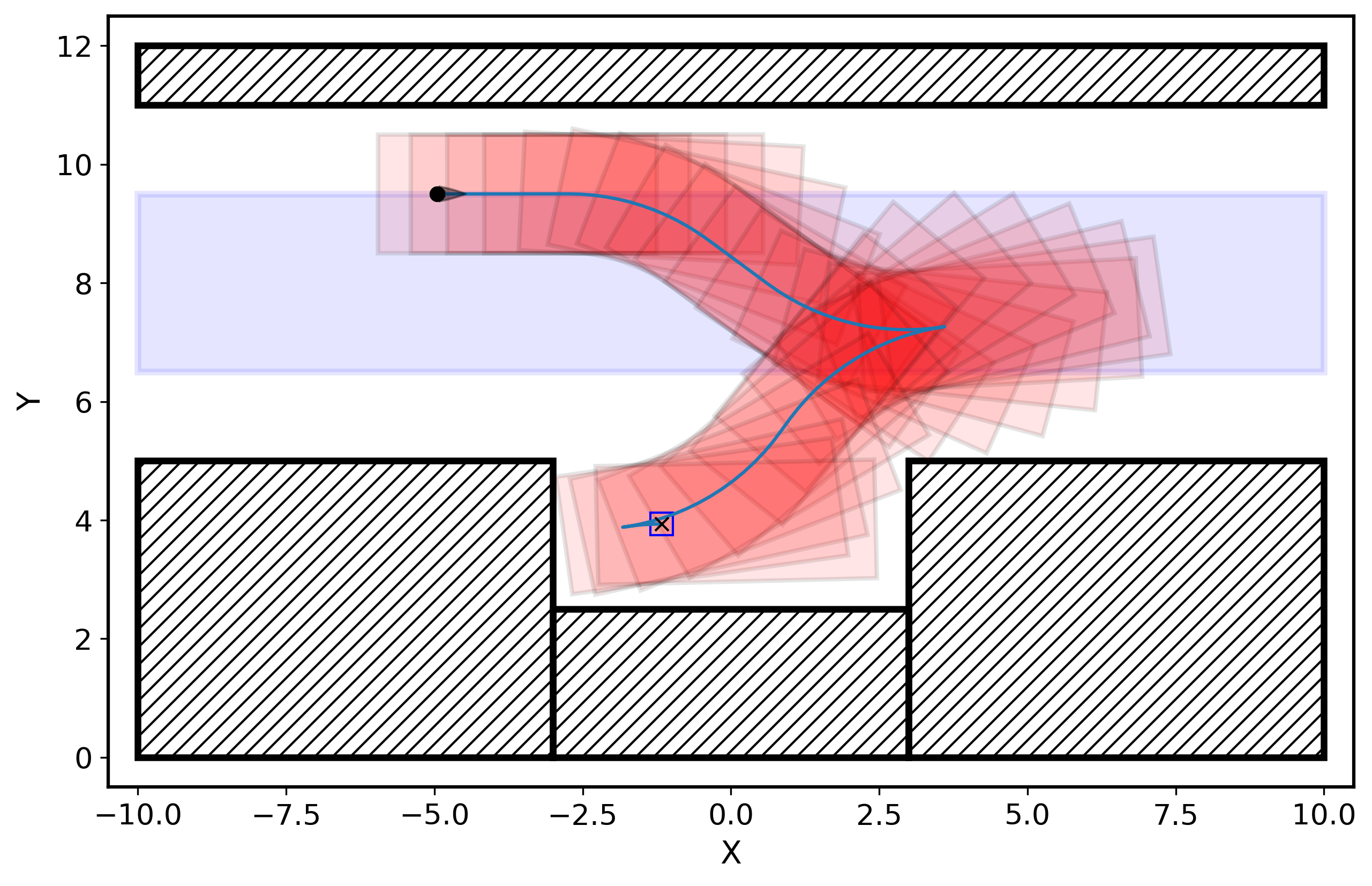}\label{fig:Parallel}}%
  \caption{Benchmark scenarios. 
  Initial states are randomly generated within a blue region defined by $\left[{c_x}_{\textup{min}},{c_x}_{\textup{max}},{c_y}_{\textup{min}},{c_y}_{\textup{max}} \right]=\left[-10,10,6.5,9.5 \right]$.
  }
  \label{fig:autonomous parking}
\end{figure}

To evaluate the effectiveness of our GeoPro algorithm, we conducted benchmarks comparing it with the state-of-the-art Optimization-Based Collision Avoidance (OBCA) method~\cite{zhang2020optimization}.
We focused on two commonly encountered scenarios: parallel and vertical parking, as depicted in Fig.~\ref{fig:autonomous parking}.
The vehicle model is given by
\begin{equation}
    \dot{c}_x=v\cos{\theta}, \dot{c}_y=v\sin{\theta},
    \dot{\theta} = \frac{v}{L}\tan{\delta},
    \dot{v}=a,
\end{equation}
where $L=2.7$ m is the wheelbase length. 
The system states are $\boldsymbol{x}=\left[ c_x, c_y, \theta, v\right]$ with control inputs $\boldsymbol{u}=\left[\delta, a \right]$.

Given the nonlinear and non-convex nature of the problem, a hybrid A* algorithm is employed to generate initial guesses for the optimization. 
All common parameters, including constraints and the ego vehicle dimensions, are set to the same for fairness.
The results are summarized in Tab.~\ref{tab:benchmark}. 
In OBCA, dual variables are utilized to reformulate the signed distance function, second-order information of cost function and constraints are required during the iterative process. 
Although OBCA employs the IPOPT solver, which is implemented in C++, the GeoPro-based approach demonstrates greater computational efficiency.
This increased efficiency is attributable to GeoPro's ability to handle non-smooth safety constraints, as well as its requirement for only first-order information.

\begin{table}[t!]
    \centering
    \caption{The performance comparison of our approach GeoPro against OBCA in 100 random tests. }
    \label{tab:benchmark}
    \resizebox{0.48\textwidth}{!}{%
\begin{tabular}{|c|c|c|c|c|c|}
\hline
\multicolumn{1}{|l|}{Scenarios}   & Algorithm                                              & mean            & std             & min             & max             \\ \hline
\multirow{5}{*}{ Vertical Parking} & \begin{tabular}[c]{@{}c@{}}OBCA\\ IPOPT\end{tabular}   & 655 ms          & 167 ms          & 380 ms          & 1027 ms         \\ \cline{2-6} 
                                  & \begin{tabular}[c]{@{}c@{}}GeoPro\\ ALSPG\end{tabular}   & \textbf{222 ms} & \textbf{130 ms} & \textbf{52 ms}  & \textbf{885 ms} \\ \cline{2-6} 
                                  & \begin{tabular}[c]{@{}c@{}}GeoPro\\ SLSQP\end{tabular} & 837 ms          & 221 ms          & 204 ms          & 1319 ms         \\ \hline
\multirow{5}{*}{ Parallel Parking} & \begin{tabular}[c]{@{}c@{}}OBCA\\ IPOPT\end{tabular}   & 708 ms          & 199 ms          & 368 ms          & 1346 ms         \\ \cline{2-6} 
                                  & \begin{tabular}[c]{@{}c@{}}GeoPro\\ ALSPG\end{tabular}   & \textbf{470 ms} & \textbf{136 ms} & \textbf{199 ms} & \textbf{909 ms} \\ \cline{2-6} 
                                  & \begin{tabular}[c]{@{}c@{}}GeoPro\\ SLSQP\end{tabular} & 942 ms          & 249 ms          & 227 ms          & 1894 ms         \\ \hline
\end{tabular}}
\end{table}

\subsection{Franka Emika Experiments}
We finally demonstrate GeoPro on a Franka Emika robot for an insertion task.
The task is to insert an polytope into a hole while avoiding four polytopic obstacles as illustrated in Fig.~\ref{fig:cover_figure}.
The behavior is composed of a non-conservative safe behavior among four rectangular  obstacles and a goal reaching behavior requires alignment of the polytope peg with the hole and mandates that the robot adhere to zero-velocity constraints upon completion of the insertion.
\section{Conclusion}
We present a geometric constrained based optimization framework that is flexible, efficient and versatile to design robot behaviors from low-dimensional to high-dimensional robot dynamics.
We successfully demonstrate this in extensive simulations and an insertion task on a Franka Emika robot arm.

We wish to extend and improve our work by the following points.
First, we intend to extend $\mathcal{C}_i$ to an arbitrary nonlinear and non-convex set.
The off-the-shelf proximal methods will be employed to solve the subproblems.
Second, safety-aware projection will be fully studied and compared against the popular CBFs based approaches.
Third, the GeoPro based constrained optimization depends on the performance of GeoPro. 
We would like to introduce Geometric Algebra to design more efficient and versatile GeoPro.

\clearpage
\bibliographystyle{IEEEtran}
\bibliography{main.bib}
\clearpage
\appendix
\section{appendix}
We reformulate the optimization problem~\ref{OCP} by classic augmented Lagrangian method.
Then we recall some results in spectral projected gradient descent method which will be used to solve GeoPro~ based constrained optimization problems~(\ref{eq:GeoPro_unconstrained}).
Additional analysis for the GeoPro and the algorithm is provided to further explain the pros and cons of our algorithm.
\subsection{GeoPro based Constrained Optimization}
In this section, we introduce the classic augmented Lagrangian approach for the solution of problem $\ref{OCP}$.
The augmented Lagrangian is formulated as: 
\begin{equation}
\mathcal{L}(\boldsymbol{x},\boldsymbol{u}, \boldsymbol{\lambda}, \boldsymbol{\rho}):=c(\boldsymbol{x}, \boldsymbol{u})+\sum^{N_p}_{i=1} \frac{\rho_{\mathcal{C}_i}}{2} d_{\mathcal{C}_i}^2\left(g_i(\boldsymbol{x})+\frac{\boldsymbol{\lambda}_{\mathcal{C}_i}}{\rho_{\mathcal{C}_i}}\right),
\end{equation}
where $\boldsymbol{\lambda} = \left[ \boldsymbol{\lambda}_{\mathcal{C}_1}, \ldots, \boldsymbol{\lambda}_{\mathcal{C}_{N_p}}\right]$, $\boldsymbol{\lambda}_{\mathcal{C}_i} \in \mathbb{R}^{N n_i}$ is the Lagrangian multipliers.
$\boldsymbol{\rho}=\left[ \rho_{\mathcal{C}_{1}}, \ldots, \rho_{\mathcal{C}_{N_p}}\right]$, $\rho_{\mathcal{C}_i} \in \mathbb{R}$ is the penalty parameters.
Incorporating GeoPro~\ref{GeoPro} yields the following compact form for the augmented Lagrangian:
{\footnotesize
\begin{equation}
\mathcal{L}(\boldsymbol{x},\boldsymbol{u},\boldsymbol{\lambda}, \boldsymbol{\rho}):=c(\boldsymbol{x}, \boldsymbol{u})+\sum^{N_p}_{i=1} \frac{\rho_{\mathcal{C}_i}}{2}||g_i(\boldsymbol{x})+\frac{\boldsymbol{\lambda}_{\mathcal{C}_i}}{\rho_{\mathcal{C}_i}} - \mathcal{P}^{(g_i(\boldsymbol{x})+\frac{\boldsymbol{\lambda}_{\mathcal{C}_i}}{\rho_{\mathcal{C}_i}})}_{\mathcal{C}_i} ||^2.
\end{equation}
}

To update the penalty parameters and convenience, We define a distance function as follows:
\begin{equation}\label{eq:V_compute}
    V(\boldsymbol{x}, \boldsymbol{\lambda}, \rho):=g_i(\boldsymbol{x})-\mathcal{P}^{\left(g_i(\boldsymbol{x})+ \frac{\boldsymbol{\lambda}_{\mathcal{C}_i}}{\rho_{\mathcal{C}_i}} \right)}_{\mathcal{C}_i}
\end{equation}
The norm of $V$ can be used to design a termination criterion, $\|V\| \leq \varepsilon_{\textup{tol}}$, $\varepsilon_{\textup{tol}} >0 $. 
Since there is an extra system constraint in \ref{OCP}, we can reformulate the Lagrangian and eliminate the variable of $\boldsymbol{x}$ by $\boldsymbol{x}_{k+1}=f(\boldsymbol{x}_k, \boldsymbol{u}_k)$.
Recall that $\boldsymbol{u}=\left [ \boldsymbol{u}_0;\ldots;\boldsymbol{u}_{N-1}\right ]$ and $\boldsymbol{x}=\left[\boldsymbol{x}_0;\ldots;\boldsymbol{x}_N \right]$. 
Through the standard linearization of the nonlinear system dynamics, we have:
\begin{equation}
    A_k:=\frac{\partial f}{\partial \boldsymbol{x}}\left(\boldsymbol{x}_k, \boldsymbol{u}_k\right), \quad
    B_k:=\frac{\partial f}{\partial \boldsymbol{u}}\left(\boldsymbol{x}_k, \boldsymbol{u}_k\right),
\end{equation}
where $A_k \in \mathbb{R}^{n_x\times n_x}$ and $B_k \in \mathbb{R}^{n_x\times n_u}$.
The rollout of the system dynamics will be denoted by $\boldsymbol{x}=\boldsymbol{\mathcal{A}}\boldsymbol{x}_0 + \boldsymbol{\mathcal{B}}\boldsymbol{u}$:
\begin{equation}
\scalebox{0.65}{
$\boldsymbol{\mathcal{A}}=\left[\begin{array}{c}
A_0 \\
A_1 A_0 \\
\vdots \\
\prod_{i=0}^{T-2} A_i \\
\prod_{i=0}^{T-1} A_i
\end{array}\right], 
\boldsymbol{\mathcal{B}}=\left[\begin{array}{ccccc}B_0 & 0 & \cdots & 0 & 0

\\A_1 B_0 & B_1 & \cdots & 0 & 0

\\

\vdots & \vdots & \ddots & \vdots & \vdots \\
\prod_{i=1}^{T-2} A_i B_0 &
\prod_{i=2}^{T-2} A_i B_1&
\cdots & 
B_{T-2} & 0
\\
\prod_{i=1}^{T-1} A_i B_0 & \prod_{i=2}^{T-1} A_i B_1 & \cdots & A_{T-1}B_{T-2} & B_{T-1}
\end{array}\right]$
}
\end{equation}
where $\boldsymbol{\mathcal{A}}\in \mathbb{R}^{N\times n_x\times n_x}$ and $\boldsymbol{\mathcal{B}}\in \mathbb{R}^{N\times n_x \times n_u}$.
For convenience, we use $\phi(\boldsymbol{u}):=\{\boldsymbol{x}\in\mathbb{R}^{N\times n_x}: \boldsymbol{\mathcal{A}}x_0 +\boldsymbol{\mathcal{B}}\boldsymbol{u}\}$.
Let $\boldsymbol{J}_x=\frac{\partial c(\boldsymbol{x, \boldsymbol{u}})}{\partial \boldsymbol{x}}\in \mathbb{R}^{N\cdot n_x}$ and $\boldsymbol{J}_u=\frac{\partial c(\boldsymbol{x}, \boldsymbol{u})}{\partial \boldsymbol{u}}\in \mathbb{R}^{N\cdot n_u}$ be the Jacobian matrix of the cost function.
The cost function can be rewritten as:
$c(\boldsymbol{x}, \boldsymbol{u}):=c(\phi(\boldsymbol{u}), \boldsymbol{u})$.
The new Lagrangian will be: 
{\footnotesize
\begin{equation}
\mathcal{L}(\boldsymbol{u},\boldsymbol{\lambda}, \boldsymbol{\rho}):=c( \boldsymbol{u})+\sum^{N_p}_{i=1} \frac{\rho_{\mathcal{C}_i}}{2}||g_i(\boldsymbol{u})+\frac{\boldsymbol{\lambda}_{\mathcal{C}_i}}{\rho_{\mathcal{C}_i}} - \mathcal{P}^{(g_i(\boldsymbol{u})+\frac{\boldsymbol{\lambda}_{\mathcal{C}_i}}{\rho_{\mathcal{C}_i}})}_{\mathcal{C}_i} ||^2,
\end{equation}
}

The partial derivative of cost function w.r.t. $\boldsymbol{u}$ will be 
\begin{align}
    &\nabla c(\boldsymbol{u})=\frac{\partial c(\phi(\boldsymbol{u}))}{\partial \phi(\boldsymbol{u})}\frac{\partial \phi(\boldsymbol{u})}{\partial \boldsymbol{u}} + \frac{\partial c(\boldsymbol{u})}{\partial \boldsymbol{u}}=\boldsymbol{\mathcal{B}}^{\top}J_x + J_u, \\
    &\frac{\partial g_i(\boldsymbol{u})}{\partial \boldsymbol{u}}=\frac{\partial g_i(\phi(\boldsymbol{u}))}{\partial \phi(\boldsymbol{u})}\frac{\partial \phi(\boldsymbol{u})}{\partial \boldsymbol{u}}=\nabla g_i \cdot \boldsymbol{\mathcal{B}},
\end{align}

The Lagrangian derivative is:
\begin{align}
    \nabla \mathcal{L}(\boldsymbol{u}, \boldsymbol{\lambda}, \rho)&=\nabla c(\boldsymbol{u})+
    \sum_{i=1}^p \rho_{\mathcal{C}_i} (\frac{\partial g_i(\boldsymbol{u})}{\partial \boldsymbol{u}})^{\top} \left(V+\frac{\boldsymbol{\lambda}_{\mathcal{C}_i}}{\rho_{\mathcal{C}_i}} \right) \nonumber \\
    &=\boldsymbol{\mathcal{B}}^{\top}(J_x + \rho_{\mathcal{C}_i}\nabla g_i V)+J_u
\end{align}
To efficiently compute the Lagrangian derivative, we adopt a recursive iteration method.
Let $\boldsymbol{w}=\left[\boldsymbol{w}_0, \boldsymbol{w}_1, \ldots, 
\boldsymbol{w}_{N-1} \right]$ be the vector to multiply, and $\boldsymbol{z}=\left[ \boldsymbol{z}_0, \boldsymbol{z}_1, \ldots, 
\boldsymbol{z}_{N-1}\right]$ be the resulting vector.
we have $\boldsymbol{z}=\boldsymbol{\mathcal{B}}^{\top} \boldsymbol{w}$:
\begin{equation}
\scalebox{0.69}{
$
\left[\begin{array}{c}
\boldsymbol{\boldsymbol{z}}_0 \\
\boldsymbol{\boldsymbol{z}}_1 \\
\vdots \\
\boldsymbol{\boldsymbol{z}}_{N-2} \\
\boldsymbol{\boldsymbol{z}}_{N-1}
\end{array}\right]=\left[\begin{array}{ccccc}
B_0^{\top} & B_0^{\top} A_1^{\top} & \cdots & B_0^{\top}\prod_{i=1}^{N-2} A_i^{\top} & B_0^{\top}\prod_{i=1}^{N-1} A_i^{\top} \\
0 & B_1^{\top} & \cdots & B_1^{\top}\prod_{i=2}^{N-2} A_i^{\top} & B_1^{\top}\prod_{i=2}^{N-1} A_i^{\top} \\
\vdots & \vdots & \ddots  & \vdots & \vdots \\
0 & 0 & \cdots & B_{N-2}^{\top} & B_{N-2}^{\top}A_{N-1}\\
0 & 0 & \cdots & 0 & B_{N-1}^{\top}
\end{array}\right]
\left[\begin{array}{c}
\boldsymbol{\boldsymbol{w}}_0 \\
\boldsymbol{\boldsymbol{w}}_1 \\
\vdots \\
\boldsymbol{\boldsymbol{w}}_{N-2} \\
\boldsymbol{\boldsymbol{w}}_{N-1}
\end{array}\right],
$
}
\end{equation}
By observation, we have following backward recursive equations:
\begin{align}
    &\boldsymbol{z}_{k}=\boldsymbol{B}_{k}^{\top}\tilde{\boldsymbol{z}}_{k}, \nonumber \\
    &\tilde{\boldsymbol{z}}_{k}=\boldsymbol{w}_{k}+ A^{\top}_{k+1}\tilde{z}_{k+1}, \nonumber\\
    &\tilde{\boldsymbol{z}}_{N-1} = \boldsymbol{w}_{N-1},
\end{align}
The constrained optimization problem~(\ref{OCP}) is reformulated as a constrained and non-convex subproblem:
\begin{equation}\label{eq:GeoPro_unconstrained}
    \argmin_{\boldsymbol{u}\in\mathcal{D}_u} \mathcal{L}(\boldsymbol{u}, \boldsymbol{\lambda}, \rho)
\end{equation}
where the $D_u$ can be arbitrary closed set.
We propose to solve this subproblem by spectral projected gradient descent method with the Euclidean projection $P_{\mathcal{D}_u}(\boldsymbol{u})$~(\ref{eq: self}).

\begin{algorithm}[ht]
\SetKwInOut{KwOut}{Return}
\caption{GeoPro-based Optimization}
\label{alg:GeoPro_OCP}
\KwIn{$\boldsymbol{x}_0, \boldsymbol{u}_{\textup{init}}, \mathcal{P}:\mathcal{B}_{\textup{safe}},\mathcal{B}_{\textup{reach}},\mathcal{B}_{\textup{limit}}, \boldsymbol{\lambda}_{\mathcal{C}_i}=0, \rho_{\mathcal{C}_i}=0.1, \beta=5,k=0, N_{\textup{max}}=20,\varepsilon_{\textup{tol}}=10^{-4}$}
	{\While{$N_{\textup{iter}}\leq N_{\textup{max}}$}
	{$\boldsymbol{u}_{k+1} = \argmin_{\boldsymbol{u}\in \mathcal{D}_u } \mathcal{L}(\boldsymbol{u}_{k}, \boldsymbol{\lambda}^{k}_{\mathcal{C}}, \rho^k_{\mathcal{C}})~\ref{eq:GeoPro_unconstrained}$\par
        \For{$\mathcal{P}_{\mathcal{C}_i}$}{$\boldsymbol{\lambda}^{k+1}_{\mathcal{C}_i}=\rho^{k}_{\mathcal{C}_i}(V_{k}(\boldsymbol{u}_{k}, \boldsymbol{\lambda}_{\mathcal{C}_i}^{k}, \rho^{k}_{\mathcal{C}_i})\ref{eq:V_compute}+\frac{\boldsymbol{\lambda}_{\mathcal{C}_i}^{k}}{\rho^{k}_{\mathcal{C}_i}})$\par
        \If{$|V|_{k+1}\leq |V|_k$}{
                $\rho^{k+1}_{\mathcal{C}_i}=\rho^{k}_{\mathcal{C}_i}$
            }
        \Else{$\rho^{k+1}_{\mathcal{C}_i}=\beta\rho^{k}_{\mathcal{C}_i}$}
        }
        \If{termination condition: $|V_{k+1}| \leq \varepsilon_{\textup{tol}}$}{break}
        $N_{\textup{iter}++}$
	}
 }
 \KwOut{$\boldsymbol{u}_{\textup{opt}}$}
\end{algorithm}

\subsection{Projection-based Constrained Optimization}
Given a smooth function $f(\cdot): \mathbb{R}^{n}\rightarrow \mathbb{R}$, the idea of Cauchy's steepest descent algorithm is $\boldsymbol{x}_{k+1}=\boldsymbol{x}_k + \alpha_k d_k$, where $d_k = -\nabla f\left(\boldsymbol{x}_k\right)$ is the search direction and $\alpha_k$ is the steplength. 
A common way to choose steplength is $\min _{\alpha \geq 0} f\left(\boldsymbol{x}_k-\alpha \nabla f\left(\boldsymbol{x}_k\right)\right)$.
The projected gradient decent builds on top of a projection function to solve constrained optimization problems as 
\begin{equation}
    \text { minimize } f(\boldsymbol{x}) \text { subject to } \quad \boldsymbol{x} \in \mathcal{D}_{\boldsymbol{x}},
\end{equation}
The iteration direction of projected gradient descent is $d_{k}=\mathcal{P}_{\mathcal{D}_{\boldsymbol{x}}}(\boldsymbol{x}_{k} - \alpha_{k}\nabla f(\boldsymbol{x}_k) - \boldsymbol{x}_k$.

\subsubsection{Spectral projected gradient descent}
The convergence rate of projected gradient is disastrous if the steplength is chosen improperly which leads to spectral projected descent with non-monotone line search.
The iteration direction of SPG is $d_{k}=\mathcal{P}(\boldsymbol{x}_{k} - \gamma_{k}\nabla f(\boldsymbol{x}_k)) - \boldsymbol{x}_k$ where the $\gamma_k$ is the spectral choice of steplength.
The spectral steplength is a first order approximation of the Hessian matrix by $\gamma_{k+1} \boldsymbol{I}$ by 
\begin{equation}\label{eq:gamma}
\gamma_{k+1}^{(1)}=\frac{\boldsymbol{s}_k^{\top} \boldsymbol{s}_k}{\boldsymbol{s}_k^{\top} \boldsymbol{y}_k} \quad \text { and } \quad \gamma_{k+1}^{(2)}=\frac{\boldsymbol{s}_k^{\top} \boldsymbol{y}_k}{\boldsymbol{y}_k^{\top} \boldsymbol{y}_k},
\end{equation}
where $\boldsymbol{s}_k=\boldsymbol{x}_k-\boldsymbol{x}_{k-1}$ and $\boldsymbol{y}_k=\nabla f\left(\boldsymbol{x}_k\right)-\nabla f\left(\boldsymbol{x}_{k-1}\right)$. 
Therefore, spectral projected gradient descent is also one kind of Quasi-Newton secant methods.
The secant equation is $\boldsymbol{B}_{k+1} \boldsymbol{s}_k=\boldsymbol{y}_k$.
With the approximation, we have $\boldsymbol{\gamma}_{k+1}\boldsymbol{s}_{k}=\boldsymbol{y}_k$, the solution~\ref{eq:gamma} is obtained by least-squares that minimizes $\left\|\gamma \boldsymbol{s}_k-\boldsymbol{y}_k\right\|_2^2$.
\begin{figure}[t!]
  \centering
  \subfigure[Iteration]{\includegraphics[width=0.23\textwidth]{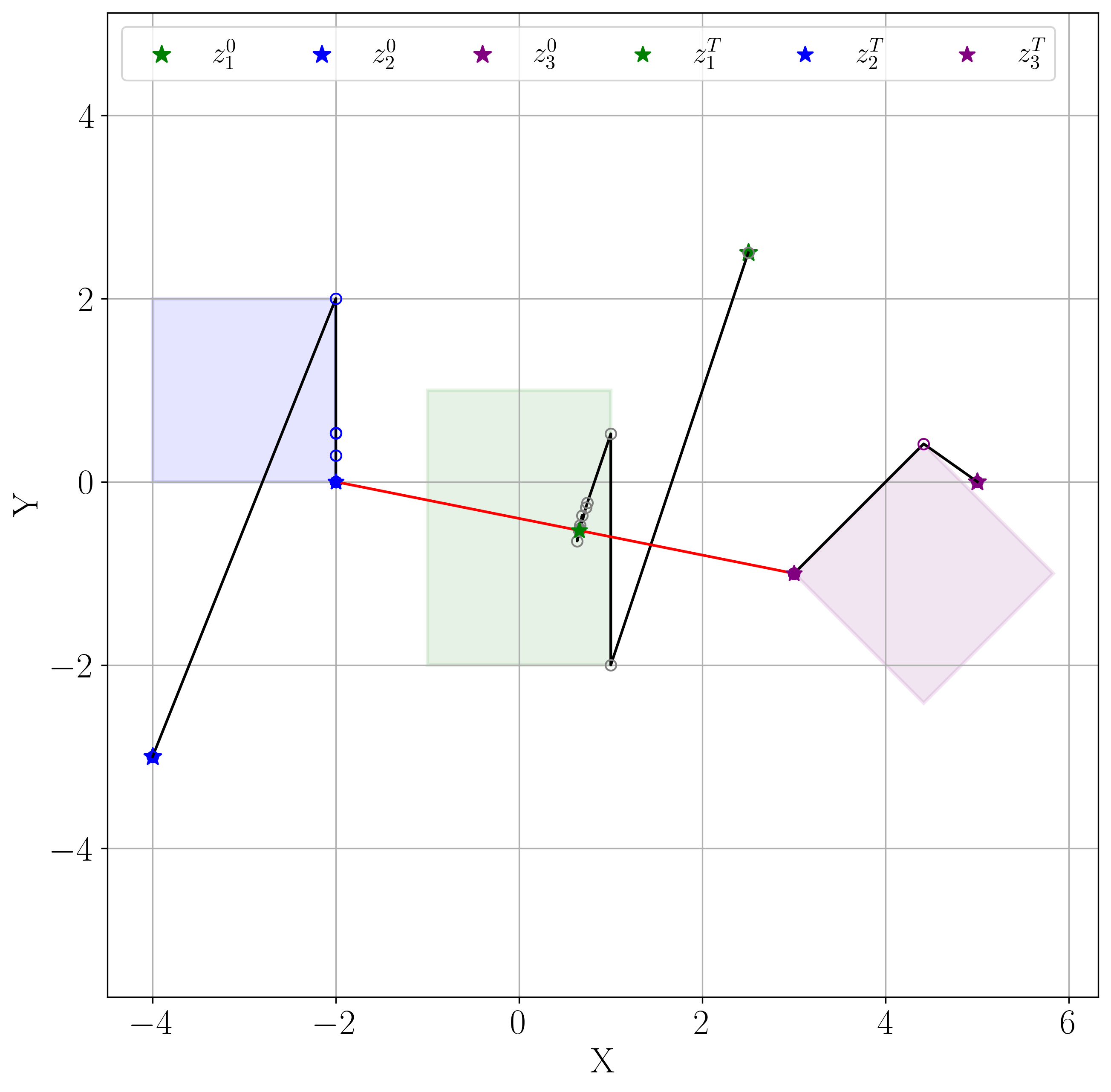}
    \label{fig:location}}
  \hfill
  \subfigure[Objective function]{\includegraphics[width=0.23\textwidth]{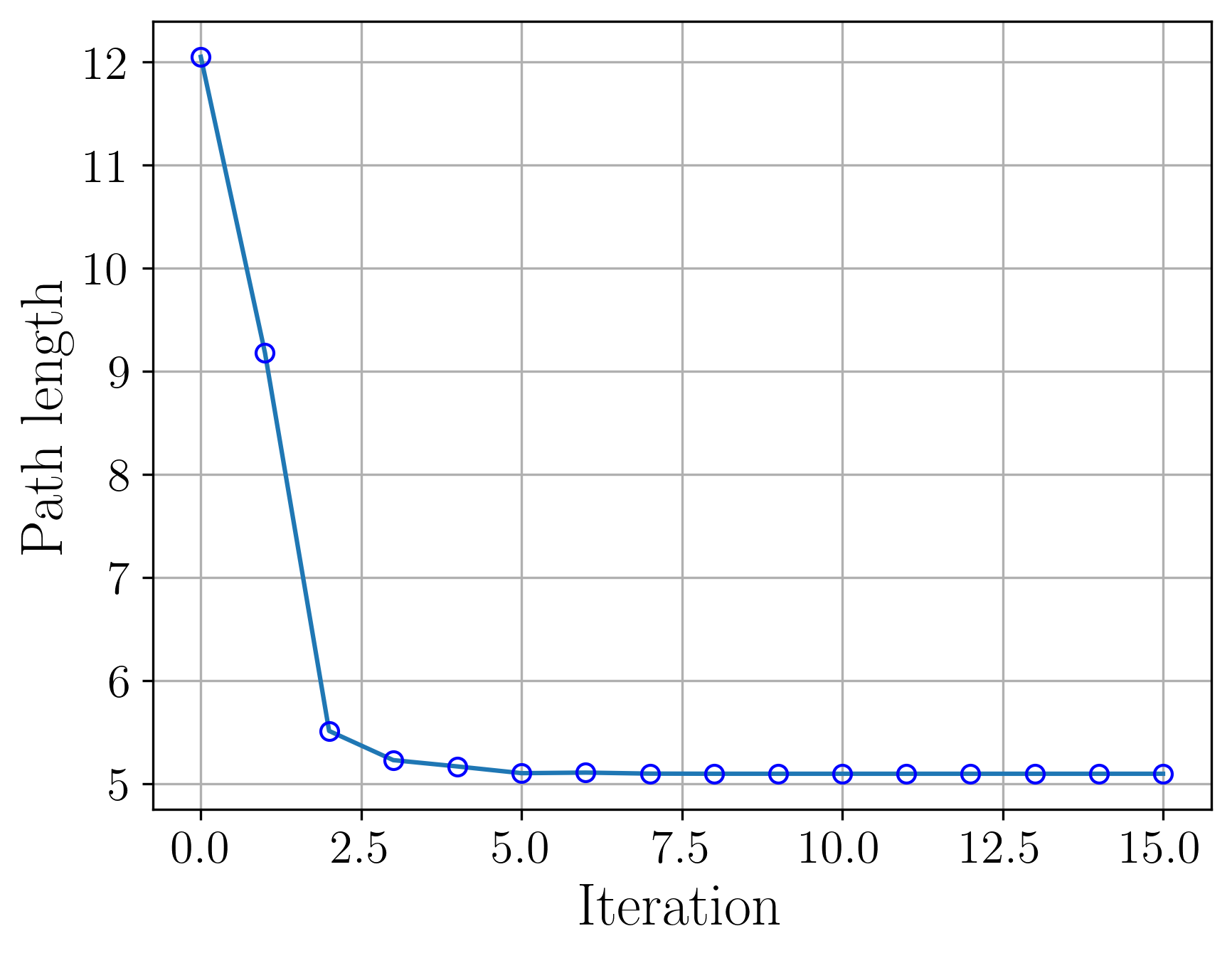}\label{fig:Objective function}}%
  \caption{The location problem.
    The task is to find the location of three variables which have the shortest path length.
    Blue, green and purple are the feasible regions for each variable respectively.}
  \label{fig:convex_spg}
\end{figure}
We demonstrate the spectral projected descent method using a 2D location problem, as shown in Fig.~\ref{fig:location}. As the variables start to iterate, the results are projected onto the feasible set after every step of the spectral gradient-based non-monotone iteration to ensure that the constraints are satisfied. Since the feasible regions are convex sets, this constitutes a convex constrained optimization problem. Non-monotone line search is employed to handle cases where the objective function is non-decreasing

\subsection{Experiments and Solver Parameters}
In this section, we present the details of setup and solver parameters used in different applications.

\subsubsection{Shape Robot Behaviors by GeoPro}
The setup parameters used in~Fig.~\ref{fig:shape behaviors} are presented here.
In (a), the obstacle positions $c_x, c_y$ are (0.15, 0.15), (-0.15, 0.15) and (-0.15, -0.15).
  In (b), the radii of the inscribed circles for five geometries are 0.5, 0.4, 0.3, 0.15, 0.1 with the same center $c_x=0.4,c_y=0$. 
  In (c), four obstacles centered at $\left[\pm 0.2, \pm 0.2 \right]$.
  The ellipse size is (0.1, 0.05) and rotated by $\frac{\pi}{4}$.
  The inscribed circles for the quadrilateral and pentagon are both 0.1.
  The box limit is $\left[-0.35, 0.35\right]$ for both $c_x, c_y$.
  The subgoals are set on a circle with radius 0.15.
  In (d), six circle centers are [0.05, -0.15], [0.15, -0.05], [0.15, -0.15], [-0.05, 0.15],[-0.15, 0.05], [-0.15, 0.15].
  The box limits are the same.
  Left circle-shaped goal has a radius of 0.05 and centered at [-0.2, 0.2].

\begin{table}[t!]
    \centering
    \caption{The implementation parameters in autonomous parking benchmark. }
    \label{tab:parameters}
\begin{tabular}{c|c|c}
\hline
Setup                              & Parameters                     & Values            \\ \hline
\multirow{6}{*}{Hybrid A*}         & Grid size                      & 0.5 m             \\
                                   & Yaw resolution                 & 0.5               \\
                                   & Motion resolution              & 0.1 m             \\
                                   & Back penalty                   & 0.5               \\
                                   & Steer change penalty           & 0.5               \\
                                   & Steer penalty                  & 0.5               \\ \hline
\multirow{4}{*}{Vehicle}           & Width                          & 2.0 m             \\
                                   & Length                         & 4.7 m             \\
                                   & Whelbase                       & 2.7 m             \\
                                   & Steering angle limit           & 0.6 rad           \\ \hline
\multirow{2}{*}{Parking spot size} & Perpendicular ($w$, $d$, $h$, $\theta$) & 2.6, 5.0, 6.0, 90 \\ \cline{2-3} 
                                   & Parallel ($w$, $d$, $h$, $\theta$)      & 6.0, 2.5, 6.0, 0 \\ \hline
\end{tabular}
\end{table}

\subsubsection{Non-holonomic Mobile Robots}
Here we present details of the setup in non-holonomic mobile robots.
We consider four geometric shapes and one polytope obstacle with 8 edges.
These shapes can designed by user preference, but in our work, the specified vertices positions are
\begin{small}
\begin{itemize}
    \item Rectangle $\left[(0,-0.25),(0.5,0),(0,0.25)\right]/s$
    \item Triangle $\left[(3,-1), (3,1), (-1,1),(-1,-1)\right]/s$
    \item Quadrilateral $\left[(2,0),(0,1),(-0.5,0),(0,-1.0) \right]/s$
    \item L shape $\left[(3,-1),(3,1),(1,1),(1,3),(-1,3),(-1,-1) \right]/s$
\end{itemize}
\end{small}
where the $s$ is the scale factor to test the size of geometries.
The obstacle is a octagon with center $\left[c_x, c_y \right]=\left[0.6, 0.5 \right]$

\subsubsection{Planar Arm}
From Fig.~\ref{fig:Ma_01} to Fig.~\ref{fig:Ma_peg}, the initial joint angles are $\left[\frac{2\pi}{3},-\frac{\pi}{6}, -\frac{\pi}{4}\right]$.
From  Fig.~\ref{fig:Ma_01} to Fig.~\ref{fig:Ma_peg},
the goal pose is $\left[c_x, c_y, \theta \right]=\left[2.5, 0.5, -\frac{\pi}{4} \right]$.
In Fig.~\ref{fig:Ma_01}, the limit is $\boldsymbol{\ddot{q}} \in \left[-3, 1\right]$.
In Fig.~\ref{fig:Ma_02}, the subgoal is to follow the straight line connected to the initial position and goal position.
In Fig.~\ref{fig:Ma_03}, the center of the circle is $\left[ c_x, c_y\right]=\left[ 2.5, 2.5\right]$, the radius is $1.5$, the distance between the end-effector is $0.5$.
In Fig.~\ref{fig:Ma_03_region}, the radius of the bigger circle is $2.0$.
In Fig.~\ref{fig:Ma_peg}, the goal pose is $\left[1.5, 1.0 \frac{-\pi}{4} \right]$.
In Fig.~\ref{fig:Ma_tele}, the initial joint angles are $\left[2.1,-0.5, -1.1\right]$,
the goal pose is $\left[2.5, 0.48, 1.57 \right]$.

\subsubsection{Autonomous Parking Benchmark}
For behaviors $\mathcal{B}_{\textup{reach}}$, the vertical parking goal pose is $\boldsymbol{p}=\left[0, 1.35, \pi/2, 0\right]$, the parallel parking goal is $\boldsymbol{p}=\left[-1.2, 4, 0, 0 \right]$.
For self-limiting behaviors $\mathcal{B}_{\textup{limit}}$, 
$v\in\left[-1,2 \right]$ m/s, $\delta \in \left[-0.6, 0.6 \right]$ rad/s and $a\in \left[-1,2 \right]$ m/s.
The parameters of hybrid A* is shown in Tab.~\ref{tab:parameters}.
The $w$ is the width of the parking spot, $d$ is the depth of the parking spot,
$h$ is the width of maneuvering space.
The IPOPT solver parameters remain default.

\end{document}